\pdfoutput=1

\documentclass[11pt]{article}

\usepackage[final]{acl}

\usepackage{times}
\usepackage{latexsym}

\usepackage[T1]{fontenc}

\usepackage[utf8]{inputenc}

\usepackage{microtype}

\usepackage{inconsolata}

\usepackage{graphicx}

\usepackage{hyperref}       
\usepackage{url}            
\usepackage{booktabs}       
\usepackage{amsfonts}       
\usepackage{nicefrac}       
\usepackage{microtype}      
\usepackage{xcolor}         

\usepackage{multirow}
\usepackage{graphicx}
\usepackage{url}            
\usepackage{amsmath,bm}
\usepackage{amssymb}
\usepackage{wrapfig}
\usepackage[ruled,linesnumbered]{algorithm2e}
\usepackage{caption}
\usepackage{subcaption}
\usepackage{bbm}
\usepackage{latexsym}
\usepackage{soul}
\usepackage{tabulary,multirow,overpic}
\usepackage{balance}
\usepackage{pifont}
\usepackage{comment}
\usepackage{xcolor}         
\usepackage{tabularx}
\usepackage{colortbl}
\usepackage{enumitem}

\newcolumntype{x}[1]{>{\centering\arraybackslash}p{#1pt}}
\newcolumntype{y}[1]{>{\raggedright\arraybackslash}p{#1pt}}
\newcolumntype{z}[1]{>{\raggedleft\arraybackslash}p{#1pt}}
\newcommand{\app}{\raise.17ex\hbox{$\scriptstyle\sim$}}

\newcommand{\x}{{$\times$}}
\newcommand{\tablestyle}[2]{\setlength{\tabcolsep}{#1}\renewcommand{\arraystretch}{#2}\centering\footnotesize}
\newlength\savewidth\newcommand\shline{\noalign{\global\savewidth\arrayrulewidth
  \global\arrayrulewidth 1pt}\hline\noalign{\global\arrayrulewidth\savewidth}}

\definecolor{deemph}{gray}{0.5}
\newcommand{\gc}[1]{\textcolor{deemph}{#1}}

\definecolor{nicegreen}{RGB}{34,139,34}
\newcommand{\pacc}[1]{{\bf \fontsize{7.5}{42}\selectfont \color{nicegreen!80}~(#1)}}
\newcommand{\macc}[1]{{\bf \fontsize{7.5}{42}\selectfont \color{red!60}~(#1)}}
\newcommand{\method}{\textit{\textbf{Alt}ogether}}

\title{\textit{Alt}ogether: Image Captioning via Re-aligning Alt-text}

\author{Hu Xu$^{1}$, Po-Yao Huang$^{1}$, Xiaoqing Ellen Tan$^{1}$,\\
\textbf{Ching-Feng Yeh}$^{1}$\textbf{,} \textbf{Jacob Kahn}$^{1}$\textbf{,} \textbf{Christine Jou}$^{1}$\textbf{,}\\ 
\textbf{Gargi Ghosh}$^{1}$\textbf{,} \textbf{Omer Levy}$^{1}$\textbf{,} \textbf{Luke Zettlemoyer}$^{1,2}$\textbf{,} \textbf{Wen-tau Yih}$^{1}$\textbf{,}\\
\textbf{Shang-Wen Li}$^{1}$\textbf{,} \textbf{Saining Xie}$^{3}$ \and \textbf{Christoph Feichtenhofer}$^{1}$\\
$^1$Meta FAIR\quad $^2$University of Washington\quad $^3$New York University\\
\href{https://github.com/facebookresearch/MetaCLIP}{\color{purple}{https://github.com/facebookresearch/MetaCLIP}}\\
}

\begin{document}
\maketitle
\begin{abstract}
This paper focuses on  creating synthetic data to improve the quality of image captions.
Existing works typically have two shortcomings. First, they caption images from scratch, ignoring existing alt-text metadata, 
and second, lack transparency if the captioners' training data (e.g.~GPT) is unknown.
In this paper, we study a principled approach \method~based on the key idea to edit and \textit{re-align} existing \textit{alt-text}s associated with the images.
To generate training data, we perform human annotation where annotators start with the existing alt-text and re-align it to the image content in multiple rounds, consequently constructing captions with rich visual concepts. This differs from prior work that carries out human annotation as a one-time description task solely based on images and annotator knowledge. We train a captioner on this data that generalizes the process of re-aligning alt-texts at scale.
Our results show our \method~approach leads to richer image captions that also improve text-to-image generation and zero-shot image classification tasks.

\end{abstract}

\section{Introduction}

Human social interactions often gravitate towards engaging with individuals who exhibit a higher level of intelligence. This inherent social behavior underscores the aspiration to develop AI agents that surpass the average human intelligence. The pursuit of creating such advanced AI agents hinges significantly on the \textit{quality} of the training data, which ideally encapsulates superhuman intelligence.

However, in the context of image captioning, most existing training data is designed for naive and well-known visual concepts that provide little value to an average user, e.g., a caption ``a dog is walking in the park'' offer minimal utility to most users unless specific accessibility needs are present, e.g., for individuals with visual impairments. The primary issue with these captions lies in their lack of detail; they fail to convey nuanced information about the images, such as the breed of the dog or the specific name or location of the park.

Moreover, while alternative text (alt-text) in web-crawled data often contains detailed and concrete visual descriptions, current captioning models generally ignore this information. Instead, these models tend to generate captions solely based on the image content, which misses the opportunity to enhance the relevance and accuracy of the captions.

Additionally, advancements in caption quality often lack transparency and are not easily reproducible. For instance, recent developments such as LLaVA~\cite{liu2024visual} and ShareGPT4V~\cite{chen2023sharegpt4v} utilize high-quality captions derived from proprietary models like GPT-4V. While these models benefit from high-quality annotations, they are built on processes that are not openly shared. This lack of disclosure presents significant challenges in terms of scalability, intellectual property rights, data integrity and privacy. The use of such proprietary models in industry applications is fraught with risks, particularly when the implementation details remain undisclosed.

\begin{figure*}[!ht]
\centering
\includegraphics[width=0.9\linewidth] {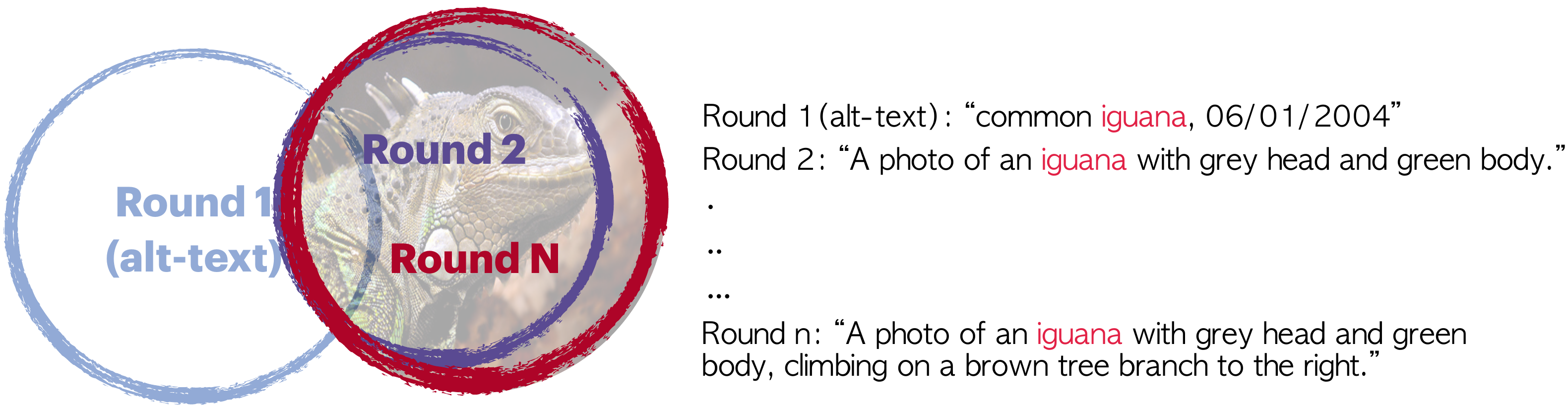}
\caption{A Venn diagram illustrating caption quality improvement via multiple rounds of \textit{re-aligning} previous captions (starting from alt-text) to the image.}
\label{fig:teaser}
\end{figure*}

This paper presents a principled approach to enhance caption quality and develops a parameter-efficient captioner capable of scaling re-captioning efforts. We assume each image contains information that the caption needs to align with using natural language. Although obtaining the real-world information from an image or generating a perfect ground-truth caption might be challenging, we demonstrate that caption \textit{quality} can be improved relatively by \textit{iteratively refining} captions to better describe the visual content (e.g., adding information on specific objects, colors, spatial relations or more fine-grained named entities).

Our key insight is that the creator who posts an image along with its associated alt-text is likely the most knowledgeable expert regarding the concrete visual concepts within that image (e.g., knowing that the animal is an "iguana" instead of just an "object," "animal," or "lizard"). It would be difficult for an average annotator to provide similar level of detail within a short annotation timeframe. Instead, these annotators could offer weak yet complementary supervision by either removing non-existent information from the alt-text or describing missing objects using more general concepts ("lizard" instead of "iguana").

Building on this insight, we introduce \method, an approach to improve image captions through the process of \textit{re-aligning existing alt-texts} with the image content. We instantiate this idea in two forms (i) through human \textit{annotation} to create a fine-tuning dataset and (ii) through a parameter-efficient \textit{captioner} that can re-caption billions of images when fine-tuned for this task.

For annotation (i), we perform multiple rounds of alt-text realignment to preserve concrete visual concepts while adding or removing relevant information, as depicted in Fig.~\ref{fig:teaser}. Starting with the initial alt-text, which may partially overlap with the image, subsequent annotation rounds iteratively refine the captions to achieve better alignment with the image's information. Using this data, we can train a captioner (ii) that is capable of generalizing this process by \textit{reading, grounding, and transforming alt-texts} into dense captions at scale. 

We evaluate our re-aligned captions across captioning,  generative and discriminative tasks. With a lightweight text decoder, our captioner surpasses alt-texts by 4\% in CLIP~\cite{radford2021learning} score and outperforms state-of-the-art captioners on a challenging test set, which we annotate based on a subset of the WIT (Wikipedia Image-Text) dataset~\cite{srinivasan2021wit}. We further evaluate our approach on text-to-image (T2I) generation, where we observe significant improvements in similarity between generated images and text prompts when training  latent diffusion models with synthetic captions.
For discriminative tasks, we obtain 1.1\% absolute accuracy improvement over 26 zero-shot classification datasets and a 3\% gain on retrieval tasks, when using synthetic captions to supplement CLIP training. An interesting observation we make is that generative and discriminative tasks require \textit{widely} different ratios (100\% vs.~15\%) of synthetic data.

\section{Related Work}
\label{sec:related_work}
\paragraph{Synthetic Data and Image Re-captioning.}
Synthetic data has recently regained popularity \cite{nguyen2024improving,li2023role} with DALL$\cdot$E~3 \cite{betker2023improving} replacing low-quality web data with synthetic data for learning image generators. Since the alt-text of web images serves various purposes and may not fully align with the images they describe, DALL$\cdot$E mixes alt-texts with synthetic captions to promote better control in image generation.
Early work ~\cite{chandu2020denoising} uses sub-selecting content words as skeletons to help generating
improved and denoised captions.
Another very recent line of concurrent research uses LLMs to fuse or combine alt-texts with captions generated from an off-the-shelf captioner~\cite{lai2024veclip,yu2024capsfusion}. However, the fusion is in language space only and has no access to the image for alignment. The resulting text may include information not present in the image and the fusion behavior of the LLM is unknown for alt-texts. See Table~\ref{tbl:eval_captioner_imagenet} for potential issues of not using vision information.

\paragraph{Dense Captioning.}
While image captioning is well-studied, generating dense captions precisely aligned with the original images has gained more attention recently. MSCOCO-style captions \cite{lin2014microsoft} are brief and describe main objects, limiting their value for aligned image-text pairs due to their brevity, general concepts, and constrained image distribution. The DCI dataset \cite{urbanek2023picture} overcomes the brevity issue but still suffers from the other limitations. DOCCI~\cite{onoe2024docci} and ImageInWords (IIW) \cite{garg2024imageinwords} address these challenges for specific datasets using clustering or iterative refinement with object detection tools. Our work proposes a general process to improve caption quality for web images, paving the way for further advancements in this area.
\paragraph{Retrieval Augmented Generation.}
Realigning alt-texts inherently grounds the captioner on input alt-texts, which is analogous to Retrieval Augmented Generation (RAG)~\cite{lewis2020retrieval,gao2023retrieval} in terms of taking additional knowledge as input.
Image captioning also adopts RAG for caption generation ~\cite{ramos2023smallcap,yang2023re}.
Our captioner shares similar advantages, such as a parameter-efficient, lightweight model for training and inference at scale, reduced factoid hallucination, and updating knowledge at inference time unavailable during training.
\paragraph{Human Preference Alignment.}
Image captioning, as an alignment problem between captions and corresponding images, relates to alignment for human preference~\cite{ouyang2022training}. However, image captioning alignment is more objective due to the clear target of aligning with information present in the image, whereas human preference alignment is subjective, as preferences can be undefined and vary among individuals.

\section{\textit{Alt}ogether: Re-aligning Alt-texts}
\label{sec:method}
This section presents our method for re-aligning alt-texts to produce dense captions with concrete visual concepts, which we later (\S\ref{captoner_implementation}) instantiate in a parameter-efficient captioner scalable to billions of images. We structure this section into three main parts: (\S\ref{sec:dalle3}) revisiting the image captioning task, (\S\ref{sec:rescap}) incorporating re-alignment into existing captioning frameworks, as well as designing annotation tasks (\S\ref{sec:annotation}) and learning mechanisms (\S\ref{sec:captioner}) for re-aligning alt-texts.

\subsection{Image Captioning}
\label{sec:dalle3}
We formulate image captioning by predicting caption tokens conditioned on the latent space of an image embedding. The loss function is defined as:
\begin{equation}
L(t, i) = \sum_j \log P(t_j |t_{j-k}, \dots, t_{j-1}; F(i); \Theta),
\label{eq:captioning}
\end{equation}
where $i$ represents an image, $F(i)$ its encoding (e.g., CLIP), $t_{j-k:j-1}$ the preceding caption tokens, and $\Theta$ the parameters of the captioner. The process involves encoding the image into a latent space and sequentially decoding the caption tokens.

\subsection{Re-aligning Previous Captions}
\label{sec:rescap}
To enhance caption accuracy, we condition the captioner on previous captions (e.g., alt-texts), 
\begin{multline}
L(t, t', i) = \sum_j \\
\log P(t_j |t_{j-k}, \dots, t_{j-1}; t'_{1:m}; F(i); \Theta),
\end{multline}
where $t'_{1:m}$ are tokens from the previous caption. This re-alignment aims to refine and better align $t'$ with the image content $i$.
\subsubsection{Annotation}
\label{sec:annotation}
We improve caption quality through \textit{iterative} human annotation, \textit{refining} previous captions (alt-texts) in multiple rounds. Starting with an initial alt-text as caption $t$, the next round uses:
\begin{equation}
t'\leftarrow t.
\end{equation}
This iterative process is designed based on the following observations: (i) the creator of alt-texts is possibly the best expert/annotator who can describe the image in fine-grained visual concepts, and it could be very challenging later for an annotator to understand and caption the image at that detail (e.g., identify and specify ``iguana'' in the caption); (ii) it is also challenging for an annotator to write a detailed caption from scratch, compared to starting from existing information. 

In experiments, we show that this iterative process of re-aligning improves the annotated data, captioner, and downstream performance after different rounds of annotation.

\begin{figure*}[!ht]
    \centering
    \vspace{-10pt}
    \includegraphics[width=0.99\linewidth] {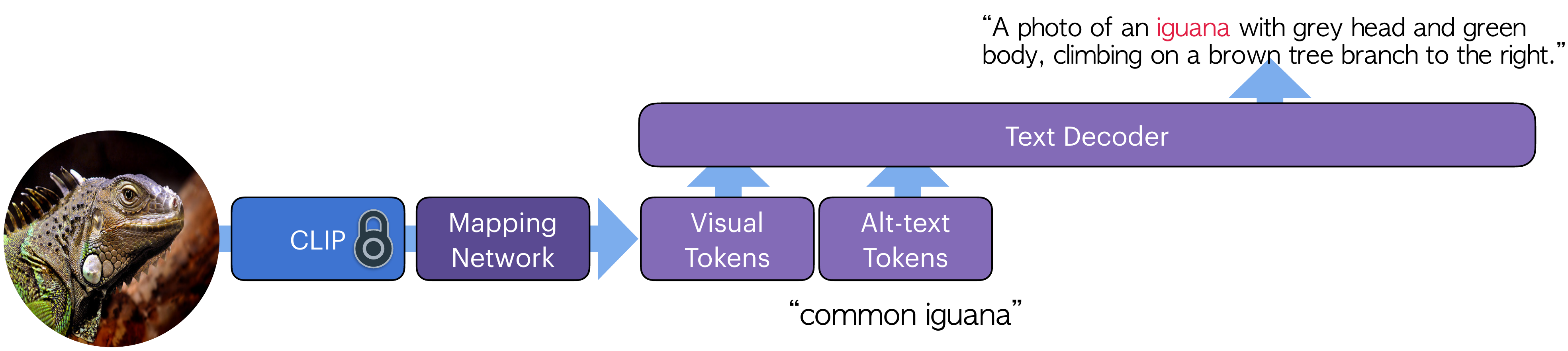}
    \vspace{-10pt}
    \caption{Re-aligning alt-texts: Our captioner takes visual \textit{and alt-text input}. We extract frozen CLIP image embeddings and transform it into a fixed number of visual tokens. Given alt-text, the decoder is able to ground this information, e.g. carrying concrete visual concepts, to generate a better caption that is aligned with the image.
    }
    \label{fig:main}
\end{figure*}

\subsubsection{Learning}
\label{sec:captioner}

We design a captioner to learn the process of \textit{re-aligning} alt-texts. We build on a simple prefix language model, ClipCap~\cite{mokady2021clipcap}, that connects a CLIP encoder and a text decoder via mapping network to implement eq.~\eqref{eq:captioning}, see Fig.~\ref{fig:main}.

\paragraph{Mapping Network.} The mapping network is a Transformer taking CLIP embeddings as input and produces visual tokens of fixed length (40 is default) that can be fed into a text decoder as the ``image prompt''.

\paragraph{Re-aligning Alt-Texts.} To model inputs on alt-texts, we simply append $m$ tokens from alt-texts, after the visual tokens. The training loss is only computed on tokens from generated captions, excluding tokens from both visual and alt-text tokens, as shown in Fig.~\ref{fig:main}. Note the alt-texts can be empty strings when these are not available.

\section{\textit{Alt}ogether: Implementation Details}
\label{captoner_implementation}
In this section, we first discuss the annotation and training data for our captioning model in \S\ref{sec:dataset}. Then we describe captioner architecture in \S\ref{sec:arch}.

\subsection{Dataset}
\label{sec:dataset}
We use a pre-training + fine-tuning framework to train the captioner, where the goal of pre-training is to learn diverse visual concepts and the later fine-tuning learns to re-align alt-texts as resulting captions.

\paragraph{Pre-training Set}
For pre-training, we randomly select 22M image-alt-text pairs from the MetaCLIP~\cite{metaclip} dataset.
This data covers long-tailed visual concepts in alt-texts which typically an average human annotator cannot infer from the image content.

\paragraph{Fine-tuning/Annotated Set.} We build a fine-tuning set (called \textit{\textbf{alt}ogether-ft}) to learn and generalize the capability of \textit{re-aligning alt-texts}.
We collect 23k images and have 3 rounds of annotation (including alt-texts as the first round). We choose two image sources: 15k images from WIT and 7k images from MetaCLIP~\cite{metaclip}.
We use these two sources to ensure rich visual concepts in alt-texts and good coverage on web images in order to mitigate the risk of inference on out-of-domain images. 
We show the annotation guidelines in Appendix \S\ref{sec:guideline} and side-by-side comparison of multiple rounds of annotation in Table \ref{tbl:sbs_annotation_wit} and Table \ref{tbl:sbs_annotation_metaclip}.

\subsection{Captioner Architecture}
\label{sec:arch}
\paragraph{Image Encoder.} We choose the pretrained \mbox{Meta CLIP} 1.0 ViT-H/14~\cite{metaclip} as the image encoder, which outputs a single embedding with 1024 dimensions. 
The image embedding is then transformed into 40 visual tokens via the mapping network to serve as the image prompt for the text decoder. 
We freeze the image encoder during the training phase and only train the mapping network.

\paragraph{Text Decoder.}
We adopt a trainable OPT 1.3B~\cite{zhang2022opt} as the text decoder for efficient training and inference (e.g., compared to Llama-13B, the throughput of this architecture is 13\x faster, see Table \ref{tbl:qps_GFLOPs}).
We append $m=128$ tokens from alt-texts after visual tokens and allow a maximum of 256 tokens for generated captions. This extends the total length of decoder to be 424 (40 visual tokens + 128 alt-text tokens + 256 generated tokens).
For alt-text tokens, we randomly sample either alt-text or empty text during training. The empty text allows the captioner to generate captions from scratch, in case the alt-texts are not available for the image.
We pre-train the captioner for 1 epoch and fine-tune on annotated data for 4 epochs.
Detailed hyperparameters are in~\S\ref{sec:hyperparams}.

\begin{table*}[t]
\centering
\vspace{-15pt}
\scalebox{0.84}{
    \setlength\tabcolsep{3.0pt}
    \begin{tabular}{l | l | l l l l | l l l}
        Captioner & CLIP Score & BLEU 1 
        & METEOR & ROUGE & CIDEr & NP F1 & NP Precision & NP Recall\\
        \toprule
        alt-text (Round 1) & 29.3 & 5.1 
        & 9.5 & 17.8 & 4.7 & 13.5 & 9.3 & 36.5\\
        \hline
        GiT & 26.3\macc{-3.0}  &0.0\macc{-5.1} 
        & 2.1\macc{-7.4} & 7.3\macc{-10.5} & 0.0\macc{-4.7} &  1.8\macc{-11.7} & 1.0\macc{-8.3} & 11.3\macc{-25.2}\\

        BLIPv2 & 28.0\macc{-1.3} & 0.2\macc{-4.9} 
        & 4.1\macc{-5.4} & 13.0\macc{-3.8} & 0.0\macc{-4.7} & 4.2\macc{-9.3} & 2.5\macc{-6.8} & 14.4\macc{-22.1} \\
        LLaVAv1.6& 27.0\macc{-2.3} & 27.7\pacc{+22.6} 
        & 10.5\pacc{+1.0} & 20.2\pacc{+2.4} & 4.9\pacc{+0.2} & 5.8\macc{-7.7} & 5.5\macc{-3.8} & 6.7\macc{-29.8} \\

        GPT-4V & 27.4\macc{-1.9} & 26.7\pacc{+21.6}
        & 10.0\pacc{+0.5} & 17.4\macc{-0.4} & 3.7\macc{-1.0} & 4.4\macc{-9.1} & 4.4\macc{-4.9} & 4.9\macc{-31.6}\\
        
        GPT-4V-turbo & 27.3\macc{-2.0} & 21.4\pacc{+16.3} 
        & 9.0\macc{-0.5} & 17.3\macc{-0.5} & 4.4\macc{-0.3} & 4.4\macc{-9.1} & 4.0\macc{-5.3} & 5.5\macc{-31.0}\\
        
        GPT-4o & 28.3\macc{-1.0} & 18.8\pacc{+13.7} 
        & 8.8\macc{-0.7} & 17.7\macc{-0.1} & 4.0\macc{-0.7} & 5.0\macc{-8.5} & 4.3\macc{-5.0} & 7.0\macc{-29.5}\\
        \hline
        \gc{\method$^{(2)}$  w/ alt} & \gc{33.1\pacc{+3.8}} & \gc{50.0\pacc{+44.9}}
        & \gc{21.5}\pacc{+12.0} & \gc{37.9}\pacc{+20.1} & \gc{48.2}\pacc{+43.5} & \gc{24.0}\pacc{+10.5} & \gc{24.1}\pacc{+14.8} & \gc{25.4}\macc{-11.1}\\ 
        
        \method$^{(3)}$ w/o alt & 32.4\pacc{+3.1} & 45.7\pacc{+40.6} 
        & 18.7\pacc{+9.2} & 34.1\pacc{+16.3} & 27.7\pacc{+23.0} & 19.2\pacc{+5.7} & 18.9\pacc{+9.6} & 20.9\macc{-15.6}\\
        \method$^{(3)}$ w/ rand alt & \gc{29.4\pacc{+0.1}} & \gc{44.6\pacc{+39.5}} & \gc{18.0\pacc{+8.5}} & \gc{33.0\pacc{+15.2}} & \gc{24.5\pacc{+19.8}} & \gc{18.7\pacc{+5.2}} & \gc{18.7\pacc{+9.4}} & \gc{20.0\macc{+16.5}}\\

        \gc{\method$^{(3)}$ w/ alt} & \gc{33.3}\pacc{+4.0} & \gc{49.6}\pacc{+44.5} 
        & \gc{21.9}\pacc{+12.4} & \gc{39.1}\pacc{+21.3} & \gc{55.6}\pacc{+50.9} & \gc{25.2}\pacc{+11.7} & \gc{24.9}\pacc{+15.6} & \gc{27.3}\macc{-9.2}\\
        \hline
    \end{tabular}
}
\caption{Evaluation of captioners on a separate test set created from the WIT dataset. We evaluate the CLIP image-text alignment score, captioning metrics which measure alignment of the model captions with \textit{ground-truth} human annotated captions: BLEU / METEOR / ROUGE / CIDEr and noun phrase (NP) F1, precision, and recall.
\method$^{(2/3)}$ indicates our captioner fine-tuned on round 2/3 annotation; `w/o alt' means captioning from scratch with no alt-text (similar to other baselines), `\gc{w/ random alt}' means captioning with randomly paired alt-texts and `\gc{w/ alt}' means captioning via re-aligning alt-texts.}

\label{tbl:text_gen}
\end{table*}

\begin{figure*}[!ht]
\centering
\includegraphics[width=0.99\linewidth] {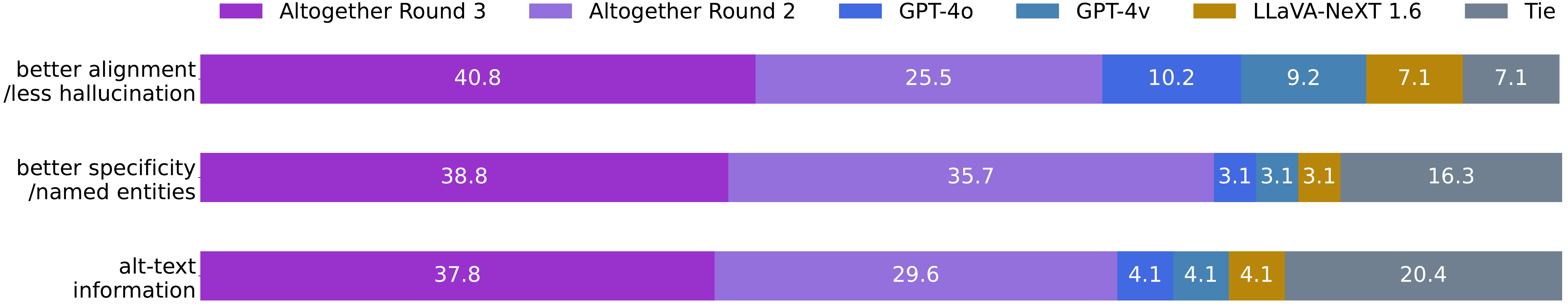}
\caption{Human evaluation on generated captions on better alignment / less hallucination (``which caption has the best alignment with the image and least hallucination''), specificity (``which caption contains more named entities'') and usefulness of alt-text information (``which caption contain most useful information from alt-texts'').}
\label{fig:human_eval}
\end{figure*}

\section{Evaluation}

Our evaluation spans three areas: (i) human annotations, (ii) captions generated from our captioner, and (iii) 
downstream tasks using our synthetic captions (i.e., text-to-image \textit{generation} and zero-shot image \textit{classification}).

\subsection{Annotated Data}
\label{sec:eval_annotated_data}
We analyze annotations in terms of length (number of words), edit distance (between annotation rounds), and CLIP image-text alignment score.
\begin{center}
\tablestyle{2pt}{1.2}	
\begin{tabular}{l|x{50}x{40}x{40}x{28}x{28}}
Annotation & Length & Edit Dist.   &  Alignment \\
\shline
Round 1 (alt-text) & 13.0  & - & 30.1 \\
Round 2 & 81.7 & 403.8 & 33.7  \\
Round 3 & 83.2 & 92.9 & 33.9  \\
\end{tabular}
\end{center}
We observe that multiple rounds of annotation (on top of the alt-text) increases the caption length and image-text alignment (CLIP score), with smaller changes in subsequent rounds. This is also reflected by the lower edit distance in the final round. We show further annotation examples in Appendix \S\ref{sec:sbs_annotation}.

\subsection{Captioner}
\label{sec:eval_captioner}

\paragraph{Human-annotated Test Set.}
We believe that existing datasets such as MSCOCO captions are not sufficient for evaluation, since these do not contain fine-grained information, e.g. a caption ``a dog sitting in a park'' does not contain information about the dog breed or park. Further, existing works ~\cite{moon2023anymal,onoe2024docci} show performance on such benchmarks correlates inversely with caption quality.
Therefore, we annotate a test set, consisting of 500 images from the WIT dataset using our 3-round annotation approach and compare our captioner to state-of-the-art captioners.
We use 3 versions of our captioner, after finetuning Round 2/3 annotations, as well as  with (w/ alt) and without (w/o alt) feeding alt-text.

We first evaluate the alignment between the images and captions via CLIP score~\cite{hessel2021clipscore} (this metric ignores the ground-truth captions and only uses CLIP similarity as metric).
The results are summarized in Table \ref{tbl:text_gen}, second column.
Our \method~ captioner improves over alt-texts by 4\% on CLIP score and significantly outperforms off-the-shelf captioners such as GiT~\cite{wang2022git,li2023blip2}, BLIP2~\cite{li2023blip2} and LLaVA~\cite{liu2024visual,liu2024llavanext}. It also outperforms proprietary captioners such as GPT-4V~\cite{GPT-4V} and GPT-4o~\cite{GPT-4o}.
The captions generated by our captioner trained with Round 3 annotation without alt-texts is worse than with alt-texts.
This implies that employing alt-texts is important for  improving image-text alignment.

Next, we compare the generated captions against the ground-truth provided by the annotators. We use BLEU/METEOR/ROUGE/CIDEr metrics and noun phrase (NP) precision, recall and F1 score. We use spaCy \url{https://spacy.io} to get two sets of NPs from generated and ground-truth captions, respectively; then we compute the intersection of these two sets as true positives. 
We observe that \method~ significantly outperforms existing captioners.
Non-dense captioners (e.g., GiT or BLIPv2) are struggling to fully describe the image with enough visual concepts (e.g., see BLIPv2's low scores across all metrics). 
\method~also outperforms dense captioners (GPT-4V/o or LLaVAv1.6), even if our model is not provided with the alt-text. 
If we provide the model with the alt-text we see a further boost in performance. This can be explained by the long-tailed visual concepts present in alt-texts~\cite{metaclip}, which is difficult for dense captioners to describe purely using image information. 

\paragraph{Low Performance of GiT and BLIPv2.}
We further investigate 0.0 CIDEr scores of GiT and BLIPv2. One reason is from using long-tailed dense captions (averaging over 80 words) as reference to compute CIDEr that penalizing short captions because CIDEr has a length penalty.
Also, both GiT and BLIPv2 are trained on the MSCOCO dataset, which typically features captions of less than 10 words focused on common objects. 
We further fine-tune GiT on \textit{\textbf{alt}ogether-ft} set for fair comparison, shown in Table \ref{tbl:git_ft}.
GiT is still far left behind \textit{\textbf{Alt}ogether}, probably because of lacking alt-texts pre-training.
Moreover, the WIT dataset includes many out-of-domain images for which these models are not optimized, leading to partial recognition issues (e.g., recognizing ``sand on the beach'' but failing to 
detail it further). Occasionally, this mismatch in training and testing also results in the generation of unreadable captions.

\begin{table}[!ht]
\centering
\scalebox{0.68}{
    \setlength\tabcolsep{3.0pt}
    \begin{tabular}{l|c|c|c|c|c}
        Baseline & CLIP Score & BLEU 1 & METEOR & ROUGE	& CIDEr\\ 
        \toprule
        GiT (MSCOCO) & 26.3 & 0.0 & 2.1 & 7.3 & 0.0\\
        GiT$^{(3)}$ w/o alt & 26.5 & 17.6 & 13.5 & 19.8 & 0.0\\
        \hline
    \end{tabular}
}
\caption{Fine-tuning GiT on \textit{\textbf{alt}ogether-ft} set.}
\label{tbl:git_ft}
\end{table}

\begin{table*}[!ht]
\footnotesize
\centering
\tablestyle{1.5pt}{1.1}
\begin{tabularx}{0.99\linewidth}{c | p{0.34\linewidth} | p{0.41\linewidth}}
Image & Alt-Texts & Generated Captions\\
\shline
\raisebox{-1\height}{\includegraphics[width=0.2\textwidth]{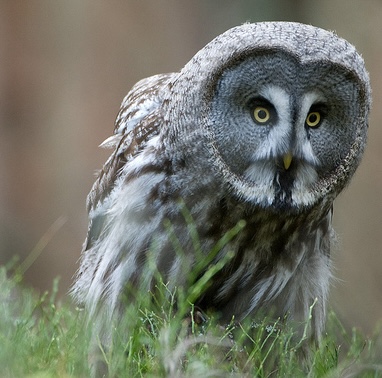}} & 
``''\newline \newline
``great gray owl, Strix nebulosa''\newline \newline
``a {\color{brown} bird}''\newline \newline
``a {\color{brown} bird} and a {\color{brown} dog}''
& A close-up photo of a {\color{red}Northern Saw-whet Owl (Aegolius nivalis)} in a natural habitat... \newline
A close-up photo of a {\color{blue}Great Gray Owl, Strix nebulosa}. The owl is standing on a grassy ...\newline
A close-up photo of a {\color{red}Northern Saw-whet Owl (Aegolius nivalis)} in a natural habitat...\newline
A close-up photo of a {\color{red}Northern Saw-whet Owl (Aegolius nivalis)} in a forest...
\\
\hline
\raisebox{-1.\height}{\includegraphics[width=0.2\textwidth]{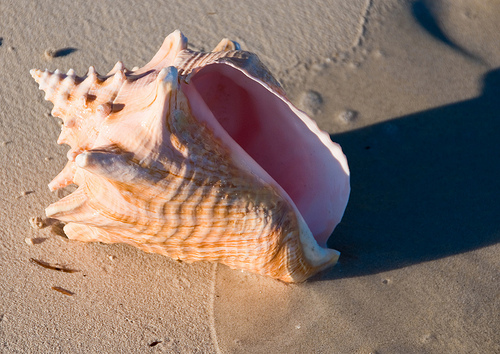}} & 
``''\newline \newline
``conch''\newline \newline
``a {\color{brown}rock}''
& 
A photo of a seashell on a sandy beach. The shell is a light pink color with ...\newline
A photo of a {\color{blue}conch} shell on a sandy beach. The shell is large and has a spiral shape... \newline
A photo of a seashell on a sandy beach. The shell is a light pink color with ...

\\
\hline
\raisebox{-1.\height}{\includegraphics[width=0.2\textwidth]{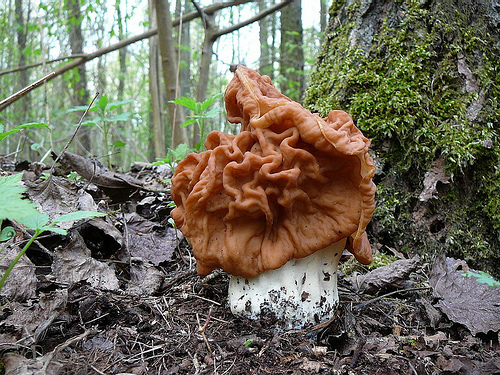}} & 
``'' \newline \newline
``gyromitra'' \newline \newline
``a {\color{brown}cat}''
& 
A photo of a mushroom, specifically a species of the genus {\color{red}Fusarium}... \newline
A close-up photo of a mushroom, specifically a species of the genus {\color{blue}Gyromitra}...\newline
A photo of a mushroom, specifically a species of the genus {\color{red}Fusarium}...
\\
\hline
\raisebox{-1.\height}{\includegraphics[width=0.2\textwidth]{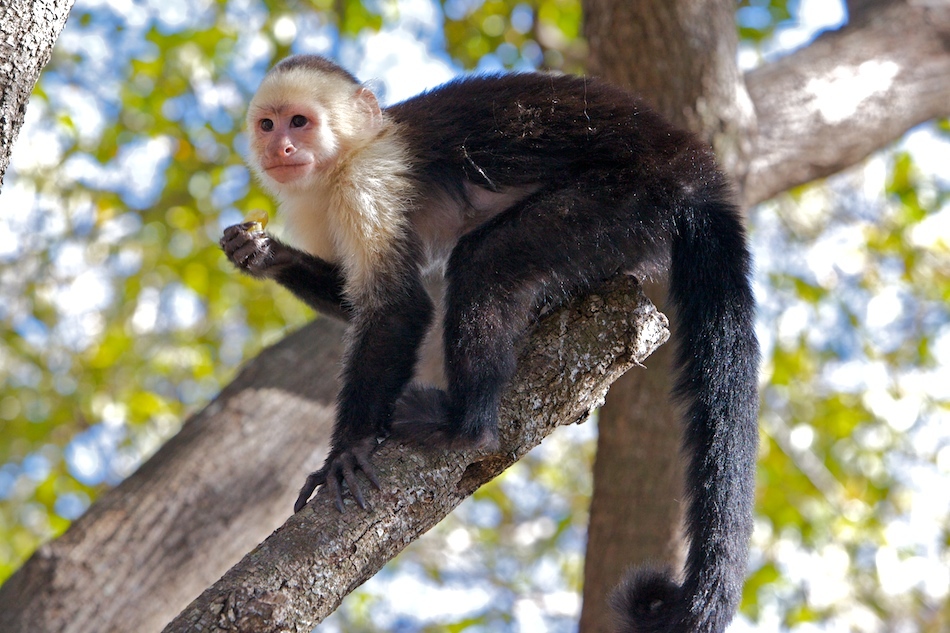}} & 
``'' \newline \newline
``spider monkey, Ateles geoffroyi'' \newline \newline
``a {\color{brown}bird}''
& 
A photo of a white-faced {\color{red}capuchin monkey (Cebus capucinus)} sitting on a tree branch... \newline
A photo of a {\color{blue}spider monkey, Ateles geoffroyi}, sitting on a tree branch. The monkey ...\newline
A photo of a white-faced {\color{red}capuchin monkey} sitting on a tree branch. The monkey has ...

\\
\hline
\end{tabularx}
\caption{Qualitative evaluation for re-aligning different alt-texts as prompts: We mark concepts carried in alt-texts in {\color{blue} blue} and erroneous captions without grounded in alt-texts in {\color{red} red}. The captioner also rejects hallucinated/general visual concepts in alt-texts in {\color{brown} brown}. This is only possible by performing alignment with text \textit{and} image information.}
\label{tbl:eval_captioner_imagenet}
\end{table*}

\paragraph{Human Study.}

We further conduct human evaluation by presenting the images, alt-texts and the captions produced by various models, and asking evaluators about three criteria: Whether the caption (i) is aligned with the image \& has fewer hallucinations; (ii) is specific (named entities, detailed description); (iii) carries useful information from the alt-text. We evaluate 5 captioners with random order when presented: LLaVAv1.6, GPT-4V, GPT-4o, and our \method~trained with Round 2/3 data. We use 3 evaluators and  100 images from WIT. The results are in Fig. \ref{fig:human_eval}. Humans highly prefer \method, and Round 3 further improves over Round 2, over the three criteria:  
\method~is also much better in (i) producing \textit{aligned} image captions \textit{without hallucination} (ii) describing images more \textit{specifically}, (iii) we see alt-texts contain useful information and captioning from scratch (LLaVA1.6, GPT-4V/o) struggles to describe this. 

To qualitatively understand the behavior of re-aligning alt-texts, we further prompt the captioner with different alt-texts on images from ImageNet, shown in Table \ref{tbl:eval_captioner_imagenet}. We try 3 different styles of alt-text prompting: (i) empty string, (ii) ImageNet class name, (iii) incorrect alt-texts. We can see that \method~can carry over concrete visual concepts and correct the hallucinated / wrong visual concepts in red that captioning from scratch (empty string) has. It further rejects alt-texts that are incorrect (e.g., alt-text ``a bird'' that is not present the image).

\begin{table*}[!ht]
\vspace{-15pt}
\centering
\tablestyle{1.5pt}{1.1}
\begin{tabularx}{0.99\linewidth}{p{0.68\linewidth}  p{0.14\linewidth}  p{0.14\linewidth} }
Prompt & Original & \method\\
\shline
\footnotesize A hummingbird {\color{red}in mid-air}, hovering above a bright red flower. The bird is mostly green with a black head and a long, pointed beak. Its {\color{red}wings are spread wide} and blurred due to the fast movement. The flower is a bright red color with {\color{red}five petals and a yellow center}. The background is a blurred green, with hints of other leaves and flowers visible.
& \raisebox{-0.8\height}{\includegraphics[width=0.13\textwidth]{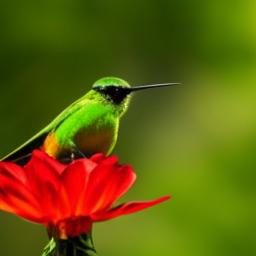}} &
\raisebox{-0.8\height}{\includegraphics[width=0.13\textwidth]{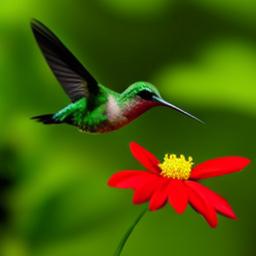}}\\
\footnotesize A {\color{red}Belgian Malinois dog} wearing {\color{red}a prosthetic leg}. The dog is standing on a grassy field with a blurred background. The prosthetic leg is made of metal and has a rubber sole. The dog is looking directly at the camera with its mouth open, as if it's smiling. {\color{red}The dog's fur is a mix of brown and black}.
& \raisebox{-0.8\height}{\includegraphics[width=0.13\textwidth]{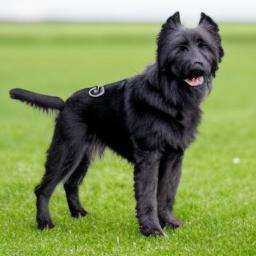}} &
\raisebox{-0.8\height}{\includegraphics[width=0.13\textwidth]{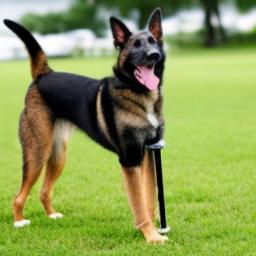}}\\
\footnotesize {\color{red}Three} potted plants, each placed in a woven rattan basket, isolated on a white background. The plants are of different sizes and species, with one being a tall, leafy plant with a thick stem, {\color{red}another being a shorter, bushy plant with a thin stem, and the third being a small, round plant with a thin stem.} The baskets are made of natural-colored wicker and have a braided design.
& \raisebox{-0.8\height}{\includegraphics[width=0.13\textwidth]{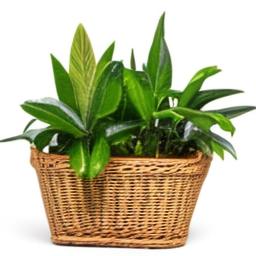}} &
\raisebox{-0.8\height}{\includegraphics[width=0.13\textwidth]{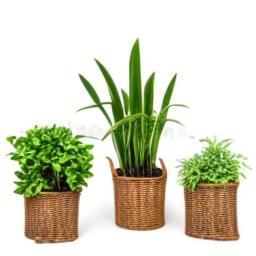}}\\
\footnotesize A beautiful, modern resort with a large swimming pool and  {\color{red}a stunning view of the sea}. The pool is surrounded by a  {\color{blue}wooden} deck with lounge chairs and  {\color{red}umbrellas}, and there are palm trees and other greenery around the pool area. In the background, you can see the  {\color{red}blue sea} and  {\color{brown}a few boats} sailing on it. The resort buildings are visible in the background, with a mix of modern and traditional architecture.
& \raisebox{-0.8\height}{\includegraphics[width=0.13\textwidth]{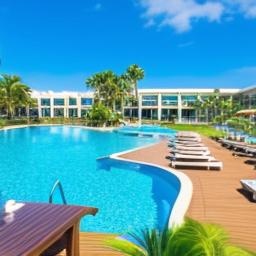}} &
\raisebox{-0.8\height}{\includegraphics[width=0.13\textwidth]{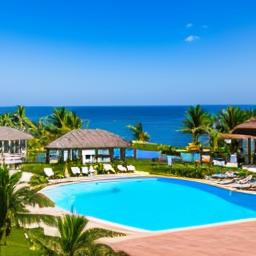}}\\
\footnotesize A scenic view of a river flowing through a forest. There is  {\color{red}a small stone bridge} with a few trees growing on either side. The bridge is made of large, rough-hewn stones and has a distinctive arched shape. The river water is clear and shallow, with a few rocks and  {\color{brown}branches} visible beneath the surface. The forest in the background is dense and green, with tall trees stretching up towards the sky.
& \raisebox{-0.8\height}{\includegraphics[width=0.13\textwidth]{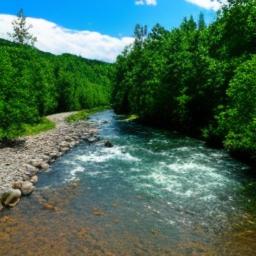}} &
\raisebox{-0.8\height}{\includegraphics[width=0.13\textwidth]{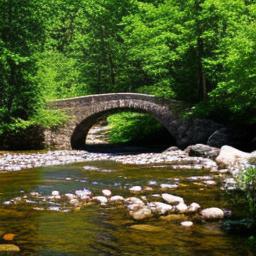}}\\
\footnotesize  {\color{red}Two tacos} on a white plate, with a violet background. Each taco has a crispy corn tortilla shell filled with shredded meat, topped with sliced avocado,  {\color{red}shredded lettuce}, and a sprinkle of red cabbage. There's a dollop of  {\color{red}creamy sauce} on top of each taco. There are  {\color{red}two glasses of drinks}, one with  {\color{brown}a pink straw and the other with a yellow straw}, placed on either side of the plate.
& \raisebox{-0.8\height}{\includegraphics[width=0.13\textwidth]{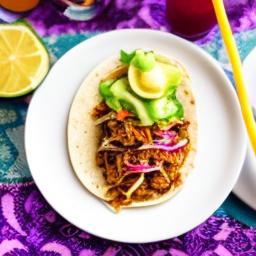}} &
\raisebox{-0.8\height}{\includegraphics[width=0.13\textwidth]{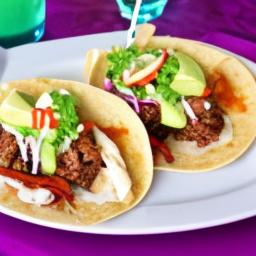}}\\
\footnotesize 
A colorful birthday cake  {\color{red}topped with a large number 9 made of fondant} and decorated with colorful sprinkles. There are also several small fondant decorations on top of the cake, including {\color{brown}a yellow chick, a pink pig, and a blue bird}. The cake is placed on a white cake stand and surrounded by colorful balloons.
& \raisebox{-0.8\height}{\includegraphics[width=0.13\textwidth]{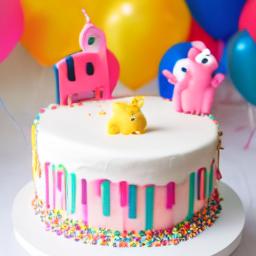}} &
\raisebox{-0.8\height}{\includegraphics[width=0.13\textwidth]{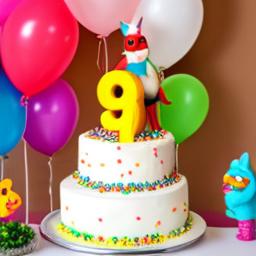}}\\
\end{tabularx}
\vspace{-5pt}
\caption{\textbf{Text-to-Image Generation}. In each group, \textit{left}: Text prompt; \textit{middle} (baseline): image generated by LDM trained with original captions; \textit{right}: image generated by LDM trained with \method~synthetic captions (Round 3). Hallucinations and errors generated by {\color{red}baseline}, {\color{blue}\method}~or {\color{brown}both} are marked with colors.
As observed, an LDM trained with \method~data follows text instruction \textit{closer} and \textit{improves image-prompt alignment} in complex scenes and specialized entities (e.g. ``a Belgian Malinois dog''). 
}
\label{tbl:qual_t2i}
\vspace{-12pt}
\end{table*}

\subsection{Text-to-image (T2I) Generation}
\label{sec:t2i}

\paragraph{Setup.}
We utilize re-aligned (synthetic) captions for training text-to-image generative models. 
Using synthetic data was shown in DALL$\cdot$E 3~\cite{betker2023improving} to be highly effective for generating images.
We use DiT-XL as the model and CC-12M~\cite{changpinyo2021cc12m} as the training dataset. 
We train the model from scratch under a controlled setup to compare the performance difference between using original captions and using re-aligned (synthetic) captions as the text inputs.
We train on CC-12M for 24 epochs on 32 A100 GPUs. Details are in Table~\ref{tbl:hp_dalle3}.

\begin{table}[h!]
\centering
\footnotesize
\vspace{10pt}
\scalebox{0.9}{
    \begin{tabular}{l | l | l }
        \multicolumn{1}{c}{}  & \multicolumn{2}{c}{Inference Prompt}\\
        \multicolumn{1}{c}{Training Data} & Original & Synthetic\\
        \toprule
        alt-texts (Round 1) & 27.0 & 28.0\\
        \method$^{(3)}$, w/o alt-texts, $p{=}0.95$ & 27.1\pacc{$+$0.1}& 29.3\pacc{$+$1.3} \\
        \hline
        \method$^{(3)}$, w/ alt-texts, $p{=}0.75$ & 27.2\pacc{$+$0.2} & 29.6\pacc{$+$1.6} \\
        \method$^{(3)}$, w/ alt-texts, $p{=}0.95$ & 27.3\pacc{$+$0.3} & 29.8\pacc{$+$1.8} \\
        \method$^{(3)}$, w/ alt-texts, $p{=}1.0$ & 27.3\pacc{$+$0.3}& 29.8\pacc{$+$1.8} \\
        \hline
    \end{tabular}
}
\caption{Evaluation of text-to-image generation on CC-12M: CLIP similarity scores between prompts (original or synthetic) and generated images.}
\label{tbl:gen_eval}
\end{table}

\newcolumntype{S}{@{}>{\lrbox0}l<{\endlrbox}}  %
\definecolor{lightgreen}{HTML}{ADD8E6}
\newcommand{\better}[1]{\colorbox{lightgreen}{#1}}
\newcommand{\datatag}[1]{\rotatebox[origin=l]{90}{\scriptsize{#1}}}
\begin{table*}[!ht]
\centering
\resizebox{1.0\linewidth}{!}{
\tablestyle{1.5pt}{1.1}
\begin{tabular}{l c | c c c c c c c c c c c c c c c c c c c c c c c c c c}
& \datatag{Average}
& \datatag{ImageNet}
& \datatag{Food-101} 
& \datatag{CIFAR10} 
& \datatag{CIFAR100} 
& \datatag{CUB}
& \datatag{SUN397}
& \datatag{Cars}
& \datatag{Aircraft}
& \datatag{DTD}
& \datatag{Pets}
& \datatag{Caltech-101}
& \datatag{Flowers}
& \datatag{MNIST}
& \datatag{FER-2013}
& \datatag{STL-10}
& \datatag{EuroSAT}
& \datatag{RESISC45}
& \datatag{GTSRB}
& \datatag{KITTI}
& \datatag{Country211}
& \datatag{PCAM}
& \datatag{UCF101}
& \datatag{Kinetics700}
& \datatag{CLEVR}
& \datatag{HatefulMemes}
& \datatag{SST2}\\
\shline
\rowcolor{lightgray} ViT-B/32 & & & & & & & & & & & & & & & & & & & & & & & & & & & \\
\method$^{(2)}$ ($p{=}1.0$) & 52.3 & 51.5 & 68.7 & 90.2 & 70.4 & 47.5 & \better{57.8} & 67.0 & 13.2 & 37.7 & 67.2 & 88.4 & \better{51.6} & 64.0 & \better{43.0} & \better{95.4} & 50.0 & \better{57.0} & 44.8 & 15.2 & 8.6 & 54.2 & 54.1 & 37.8 & \better{23.9} & 51.5 & \better{50.0}\\
\method$^{(3)}$ (w/o alt-text) & 44.5 & 39.8 & 47.4 & 88.6 & 65.7 & 14.8 & 50.0 & 54.4 & 4.9 & 29.8 & 54.2 & 79.2 & 30.4 & \better{71.9} & 25.7 & 89.6 & 39.3 & 54.2 & 37.9 & 23.9 & 5.1 & 53.5 & 47.4 & 31.5 & 15.0 & \better{54.9} & 49.2\\
\method$^{(3)}$ ($p{=}1.0$) & \better{\textbf{53.8}} & \better{52.8}  & \better{70.0} & \better{90.4} & \better{71.4} & \better{47.7} & 57.4 & \better{67.5} & \better{14.7} & \better{41.5} & \better{69.1} & \better{88.4} & 50.6 & 62.9 & 42.1 & 94.7 & \better{56.1} & 55.1 & \better{48.8} & \better{33.0} & \better{8.9} & \better{57.2} & \better{56.8} & \better{38.7} & 23.0 & 52.0 & 48.9\\
\hline
Alt-text (Round 1) & 59.3 & 68.1 & \better{84.4} & \better{93.1} & 74.5 & 66.5 & \better{67.2} & 77.9 & \better{27.9} & 59.4 & \better{90.7} & 91.7 & \better{72.0} & 25.1 & \better{45.1} & \better{97.0} & 45.8 & 63.3 & 37.0 & 30.1 & 18.8 & \better{63.3} & 67.5 & 47.7 & 19.1 & 55.9 & 52.4\\
\method$^{(2)}$ ($p{=}0.15$) & 60.3 & 67.9 & 84.1 & 92.1 & 75.3 & 66.7 & 67.1 & \better{78.2} & 25.1 & 58.8 & 89.4 & 92.5 & 70.3 & \better{37.4} & 40.2 & 95.7 & \better{55.0} & \better{67.3} & 38.3 & 31.9 & 18.0 & 59.7 & 67.4 & 48.0 & \better{33.1} & \better{56.2} & 52.9\\
\method$^{(3)}$ ($p{=}0.15$) & \better{\textbf{60.4}} & \better{68.2} & 84.3 & 92.7 & \better{75.6} & \better{67.0} & 67.1 & 77.8 & 25.6 & \better{62.6} & 89.1 & \better{92.6} & 71.2 & 36.7 & 44.5 & 96.8 & 53.2 & 63.8 & \better{38.6} & \better{35.9} & \better{18.8} & 58.2 & \better{68.1} & \better{48.2} & 24.2 & 53.5 & \better{55.1}\\
\hline
\rowcolor{lightgray} ViT-H/14 & & & & & & & & & & & & & & & & & & & & & & & & & & & \\
Meta CLIP 1.0 & 72.4 & 80.5 & 94.2 & \better{98.0} & 86.4 & 83.4 & 74.1 & 90.0 & 50.2 & 72.4 & 95.4 & 95.6 & 85.1 & 72.7 & \better{55.2} & 99.4 & 66.3 & 74.6 & \better{62.5} & \better{38.2} & 37.2 & \better{65.8} & 82.2 & 64.1 & 30.1 & \better{59.3} & \better{69.2}\\
Meta CLIP 1.2 (\method$^{(3)}$, $p{=}0.15$) & \better{\textbf{73.2}} & \better{82.1} & \better{95.0} & 97.8 & \better{87.1} & \better{88.6} & \better{74.6} & \better{93.1} & \better{63.2} & \better{73.0} & \better{95.9} & \better{95.9} & \better{86.8} & \better{86.1} & 54.6 & \better{99.5} & \better{70.3} & \better{76.0} & 57.9 & 28.1 & \better{43.3} & 50.1 & \better{85.4} & \better{65.4} & \better{32.5} & 58.3 & 62.5\\
\hline
\end{tabular}
}
\caption{Results on 26 CLIP zero-shot classification tasks. First section: Training with pure ($p{=}1.0$) synthetic captions from our captioners that were trained after different rounds of annotations. Second section: Mixing in alt-text during training (ratio of $p{=}0.15$). Third section: Comparison of a large ViT-H/14 Meta CLIP 1.2 model trained on our synthetic captions with mixed alt-text outperforms Meta CLIP 1.0~\cite{metaclip} (72.4 vs.~73.2 average accuracy).  
}
\label{tbl:clip_ablation}
\end{table*}

\begin{table}[!ht]
\centering
\resizebox{1.0\linewidth}{!}{
\tablestyle{1.5pt}{1.1}
\begin{tabular}{l |  c | c | c | c |  c }
 & Avg.~retrieval & Flickr & COCO & IN Dist. Shift & VTAB\\
\shline
\rowcolor{lightgray} ViT-B/32 & & & & &\\
Alt-text (Round 1)  & 52.6 & 72.9 & 46.6 & 52.3 & 55.3\\
\method$^{(3)}$ ($p{=}1.0$)& 46.1 & 69.0 & 42.8 & 41.7 & 47.8\\
\method$^{(3)}$ ($p{=}0.15$) & \better{\textbf{55.6}} & \better{76.0} & \better{48.9} & \better{\textbf{52.5}} & \better{\textbf{55.9}}\\
\hline
\rowcolor{lightgray} ViT-H/14 & & & & &\\
Meta CLIP 1.0 & 60.4 & 85.0 & 57.5 & 66.1 & 64.6\\
Meta CLIP 1.2 (\method$^{(3)}$, $p{=}0.15$)  & \better{\textbf{65.7}} & \better{87.6} & \better{60.7} & \better{\textbf{67.3}} & \better{\textbf{66.2}}\\
\hline
\end{tabular}
}
\caption{Zero-shot retrieval evaluation.}

\label{tbl:clip_benchmark}
\end{table}

\paragraph{Results.} We train T2I models with different mixing ratios $p$ of synthetic captions and original caption. 
During inference, following the evaluation setup in DALL$\cdot$E 3, we apply either the original prompt (alt-text) or the descriptive (synthetic) prompt as the text prompt to generate image.
We report CLIP scores to evaluate the similarities between the generated images and the corresponding text prompts on a holdout  CC-12M  set in Table~\ref{tbl:gen_eval}.

We compare T2I models trained on alt-texts (Round 1), synthetic caption with and without alt-texts grounding. 
Similar to DALL$\cdot$E 3, we first train T2I model with a high mixing ratio $p{=}0.95$ of synthetic data w/o alt-texts, mixed with original data (alt-texts).
Training with synthetic captions improve the CLIP score by 1.3\% (29.3 vs 28.0). 
Then we train a T2I model with 100\% ($p{=}1.0$) synthetic data, generated by \method~with alt-texts prompting. This yields another 0.5 gain on CLIP score. This indicates DALL$\cdot$E 3's 5\% mixing with original alt-texts is sub-optimal, not necessary and may at risk of increasing mis-aligned data, \textit{if the synthetic caption} is already \textit{re-aligned} from alt-text. Ablating  ratios of mixing existing captions (alt-text) does make a significant difference.

In Table~\ref{tbl:qual_t2i}, we qualitatively study the re-aligned captions and show this approach promotes fine-grained control and grounding for text-to-image generation with reduced hallucination.

\subsection{Classification and Retrieval}
\label{sec:clip}

\paragraph{Setup.}
Following the data curation in MetaCLIP~\cite{metaclip}, we collect 5B image-text pairs as CLIP training data. 
We follow the standard CLIP training setup for evaluating our approach using a ViT-B/32 architecture as in OpenCLIP~\cite{ilharco_gabriel_2021_5143773} and 
MetaCLIP~\cite{metaclip}.
The training hyperparameters are in Table \ref{tbl:hp_clip}.

We create 3 sets of captions by running inference on the 5B images, with captioners trained with (i) Round 2 annotation, (ii) Round 3 annotation and (iii) Round 3 without alt-texts prompts.

\paragraph{Results.}
We show the results of CLIP training by zero-shot evaluation on 26 classification tasks in Table \ref{tbl:clip_ablation}.
We first study the performance of using \textit{only} synthetic captions  (ratio of synthetic captions $p{=}1.0$). Multiple rounds of annotation help to improve accuracy by 1.5\% (Round 2 ($p{=}1.0$) vs Round 3 ($p{=}1.0$)).
Interestingly, the captioner without re-aligning alt-text (w/o alt-text) struggles (44.5\% average accuracy), indicating that re-aligning alt-text in the captioner is important. 

The next section of Table \ref{tbl:clip_ablation} shows that  training with only alt-text performs better than using only synthetic captions above. We believe this is because the captioner is likely not large enough to carry \textit{all} the alt-text information into the synthetic caption. 
We then mix alt-text and synthetic captions (ablation in Appendix \S\ref{sec:mix_clip}) for training CLIP. With a ratio of $p{=}0.15$ synthetic captions, we see a +1.1\% improvement over 26 classification tasks (Table \ref{tbl:clip_ablation}), showing how re-align can provide complementary information for CLIP training. Finally we train a large Meta CLIP 1.2 ViT-H/14 model with mixed~\method~captions and observe 73.2\% average accuracy compared to the 72.4\% with the same model in Meta CLIP 1.0~\cite{metaclip}.

Finally, we evaluate on zero-shot text-to-image retrieval tasks from DataComp~\citep{gadre2023datacomp}. Results are in Table \ref{tbl:clip_benchmark}. Mixing alt-text with synthetic captions leads to +3\% for retrieval on ViT-B and even larger gains over  Meta CLIP 1.0 ViT-H/14. 

\paragraph{Discussion.}
An interesting observation is that image generation and classification require different amount of mixing ratios for synthetic captions---the optimal mixing ratio is $\sim$100\% for T2I generation  whereas as low as $\sim$15\% for CLIP classification.
The root cause may stem from very different definitions of these two problems: T2I needs fully aligned captions to have text controlling the generated images in every detail; whereas the problem of CLIP only needs to recognize a single class name from a long-tailed vocabulary.

\section{Conclusion}
This paper presents \method, a principled way of improving image captions by re-aligning existing alt-text to images. Re-aligning alt-text allows concrete visual concepts to be carried into the resulting caption.
In experiments, we show that a lightweight captioner trained to perform this task can generate captions with significantly better captioning performance than alternatives. We further observe that the resulting captions can be used for improving both text-to-image generation and zero-shot recognition across a broad set of tasks.

\section{Limitations}
We observe the following limitations in this work:
\begin{enumerate}
    \item Evaluating captions with rare and specific concepts is challenging for the following reasons. 
    
    (i) Re-aligned alt-texts can contain superhuman information (think e.g., a very specific model type of a car or boat is not known to the majority of people). It is challenging to verify correctness, even by human evaluators. 
    
    (ii) There is no perfect metric to quantify the overall quality of alt-texts and complementary information added via re-aligning. 
    
    (iii) Lack of external high-quality ground-truth captions (that describe both alt-text and complementary information well). Note a higher quality benchmark can evaluate a lower quality caption, but not the reverse. For example, existing literature reports that benchmarks such as MSCOCO or Flicker contain only well-known visual concepts and are negatively correlated with human evaluation (IIW~\cite{garg2024imageinwords}) or higher quality benchmarks (AnyMAL~\cite{moon2023anymal}).
    \item Due to limited compute, we cannot evaluate image generation at a larger scale.
    \item Current synthetic captioning can improve alignment but cannot go beyond the concrete visual concepts described in alt-texts to improve challenging benchmarks such as ImageNet classification.
    \item 
    Working on large multimodal language models faces various constraints, including being competitive without using data from proprietary models (the community is actively distilling information from models such as GPT-4V), which leads to lack of transparency (black-box LLMs). In this work we aim to show a principled way of improving image captions with maximally preserving transparency. We will make our code, models and data available for future use.  
\end{enumerate}

\section*{Acknowledgments}
We thank Xian Li, Ping Yu, Yuandong Tian, Chunting Zhou, Armen Aghajanyan and Mary Williamson for the insightful discussion.

\bibliography{custom}

\clearpage
\appendix

\section{Annotation Guidelines}
\label{sec:guideline}
This section details our annotation guidelines. We highlight the overall goal and good practice for annotation first, then show the detailed instructions for annotators in Fig.~\ref{fig:annotation_instruct}.
Our annotations aim to enhance the \textit{alignment} between image and existing captions. We use the metadata of the image (i.e., alt-text attributes) as the starting point.
The alt-text is considered to contain ground truth information of the image but only partially describes the image.
The goal of our annotation is to significantly improve image-caption alignment 
and make the caption \textit{just} right: e.g., do not mention missing objects in the image or information beyond the image content. 

\begin{figure*}[!ht]
    \centering
    \includegraphics[width=\linewidth]{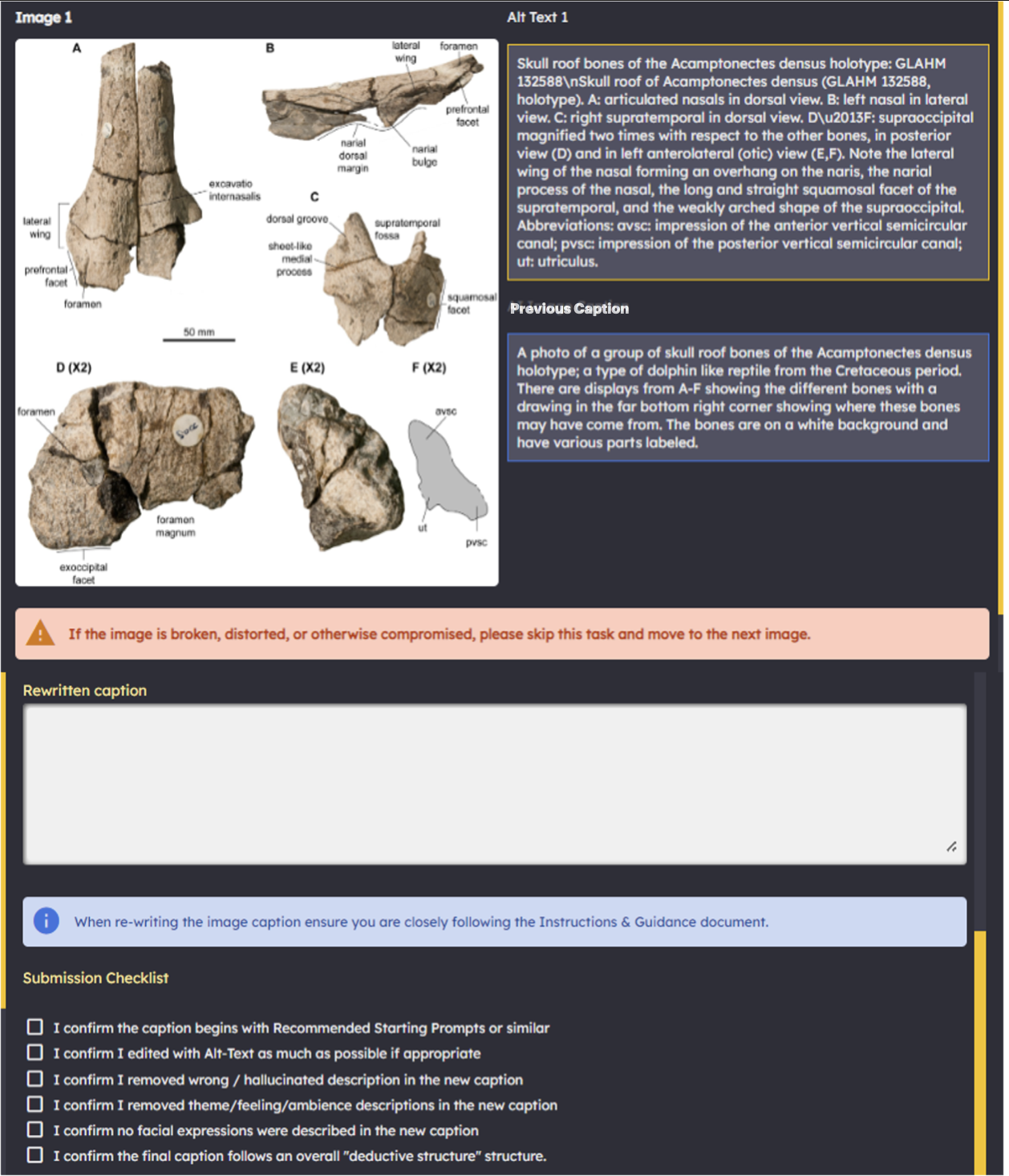} 
    \caption{Annotation interface. }
    \label{fig:annotation_ui}
\end{figure*}

\paragraph{Good Practices}
\begin{itemize}
    \item We use short prompts as the starting points of captions: such as ``a photo of", ``a painting of", ``a sculpture of", instead of verbose prompts such as ``This is an image showing ...''. We provide a recommended list of starting prompts in Table~\ref{tab:annotate_prompt}.
    \item
    We provide annotation steps to guide the annotator's workflow during annotation. See ``Annotation Steps'' in  Fig.~\ref{fig:annotation_instruct}. 
    \item We further provide a checklist to help annotators confirm if they follow each step of the guidelines well. Fig.~\ref{fig:annotation_ui} provides a screenshot of our annotation interface. 
    \item We leverage two vendors for annotation and ask each vendor to rewrite/criticise the other vendor's annotation from the previous round.
    We split the data to annotate between the two vendors, and swap the data in the next round.
\end{itemize}

{\footnotesize
\begin{table}[!ht]
\centering
\scalebox{0.9}{
\begin{tabular}{p{0.3\textwidth}}\hline
``a photo of'' \\
``a product photo of'' \\
``a low resolution photo of'' \\
``a cropped photo of''\\
``a close-up photo of''\\
``a black and white photo of''\\
``a blurry photo of''\\
``a rendering of''\\ 
``a sculpture of''\\  
``a painting of''\\  
``a cartoon of''\\
\hline
\end{tabular}}
\caption{Recommended starting prompts for captioning annotation.}
\label{tab:annotate_prompt}
\end{table}
}

\begin{table*}[!ht]
\centering
\scalebox{0.8}{
    \setlength\tabcolsep{3.0pt}
    \begin{tabular}{l | c | c | c | c}
        Decoder & Seq. Len. & Imgs per Second & GPU Days for 1B Imgs & Days on 256 GPUs for 3B Imgs\\
        \toprule
        Llama 2 13B Chat (w/o alt-texts) & 296 & 2.6 & 4452 & 52.2 \\
        \hline
        OPT 1.3B (w/o alt-texts tokens) & 296 & 19.7 & 589 & 6.8\\
        OPT 1.3B (w/ alt-texts tokens) & 424 & 15.6 & 740 & 8.6\\
        \hline
    \end{tabular}
}
\caption{Throughput of different text decoders measured on NVIDIA A100 80GB GPUs.}
\label{tbl:qps_GFLOPs}
\end{table*}

\begin{table}[!ht]
\centering
\scalebox{0.7}{
    \setlength\tabcolsep{3.0pt}
    \begin{tabular}{l|c}
        Hyperparameter & \\
        \toprule 
        Arch. & ClipCap\cite{mokady2021clipcap}\\
        Frozen Encoder & Meta CLIP 1.0 ~\cite{metaclip} \\
        Resolution & 224$\times$224 \\
        CLIP Embedding Size & 1024\\
        Visual Tokens & 40 \\
        Trainable Decoder & OPT 1.3B \\
        Attention & Flash Attention 2\\
        \hline
        Batch Size & 512 \\
        Learning Rate & 1e-3\\
        Minimal Learning Rate Ratio & 0.1\\
        Warm-up & 2k \\
        \hline
        Pre-training Data & MetaCLIP 22M\\
        Pre-training Steps & 44k\\
        Fine-tuning Data & WIT 15k + MetaCLIP 7k\\
        Fine-tuning Steps & 96 \\
        \hline
        Temperature & 0.2\\
        Top-p sampling (nucleus sampling) & 0.7\\
        \hline
    \end{tabular}
}
\vspace{5pt}
\caption{Hyperparameters of captioner training.}
\label{tbl:hp_captioner}
\end{table}

\begin{table}[!ht]
\centering
\scalebox{0.8}{
    \setlength\tabcolsep{3.0pt}
    \begin{tabular}{l|c}
        Hyperparameter & \\
        \toprule 
        Arch. & DiT-XL\\
        Activation Function & GELU\\	
        Training Data & CC12M\\
        Image Size & 256 \\
        Batch Size & 8192\\
        Learning Rate & 2.0e-5\\
        Warm-up & 1000 \\
        Training Epochs & 24 \\
        \hline
    \end{tabular}
}
\vspace{5pt}
\caption{Hyperparameters of text-to-image generation training.}
\label{tbl:hp_dalle3}
\end{table}

\begin{table}[!ht]
\centering
\scalebox{0.8}{
    \setlength\tabcolsep{3.0pt}
    \begin{tabular}{l|c|c}
        Hyperparameter & ViT-B/32 & ViT-H/14\\ 
        \toprule 
        Activation Function & QuickGELU & GELU\\	
        Seen Pairs & 12.8B & 51.2B\\
        Batch Size & 32768 & 120k\\
        Learning Rate & 5.0e-4 & 4.0e-4\\
        Warm-up & 2k & 2k\\
        \hline
    \end{tabular}
}
\vspace{5pt}
\caption{Hyperparameters of Meta CLIP 1.2 training.}
\label{tbl:hp_clip}
\end{table}

\section{Side-by-side Comparison of Multiple Rounds of Annotation}
\label{sec:sbs_annotation}
We show side-by-side comparison of annotations in Table~\ref{tbl:sbs_annotation_wit} for WIT images and Table~\ref{tbl:sbs_annotation_metaclip} for MetaCLIP images (images are not shown).

\section{Altogether Evaluated on MSCOCO}
The \textit{\textbf{Alt}ogether-ft} fine-tuning set is very different in style from the popular captioning dataset MSCOCO. As a reference, we also report performance on MSCOCO 2017 as the reference caption in Table \ref{tbl:altogether_mscoco}.

\begin{table}[!ht]
\centering
\scalebox{0.65}{
    \setlength\tabcolsep{3.0pt}
    \begin{tabular}{l|c|c|c|c|c}
        Baseline & CLIP Score & BLEU 1 & METEOR & ROUGE	& CIDEr\\ 
        \toprule 
        COCO annotation & 30.37 & - & - & - & - \\
        \method$^{(3)}$ w/o alt & 33.69 & 17.5 & 17.3 & 19.0 & 0.0\\
        \hline
    \end{tabular}
}
\vspace{5pt}
\caption{Altogether evaluated on MSCOCO.}
\label{tbl:altogether_mscoco}
\end{table}

\section{Ratio of Mixing Synthetic Captions for CLIP Training}
\label{sec:mix_clip}

\begin{figure}[!t]
\centering
\includegraphics[width=1.0\linewidth] {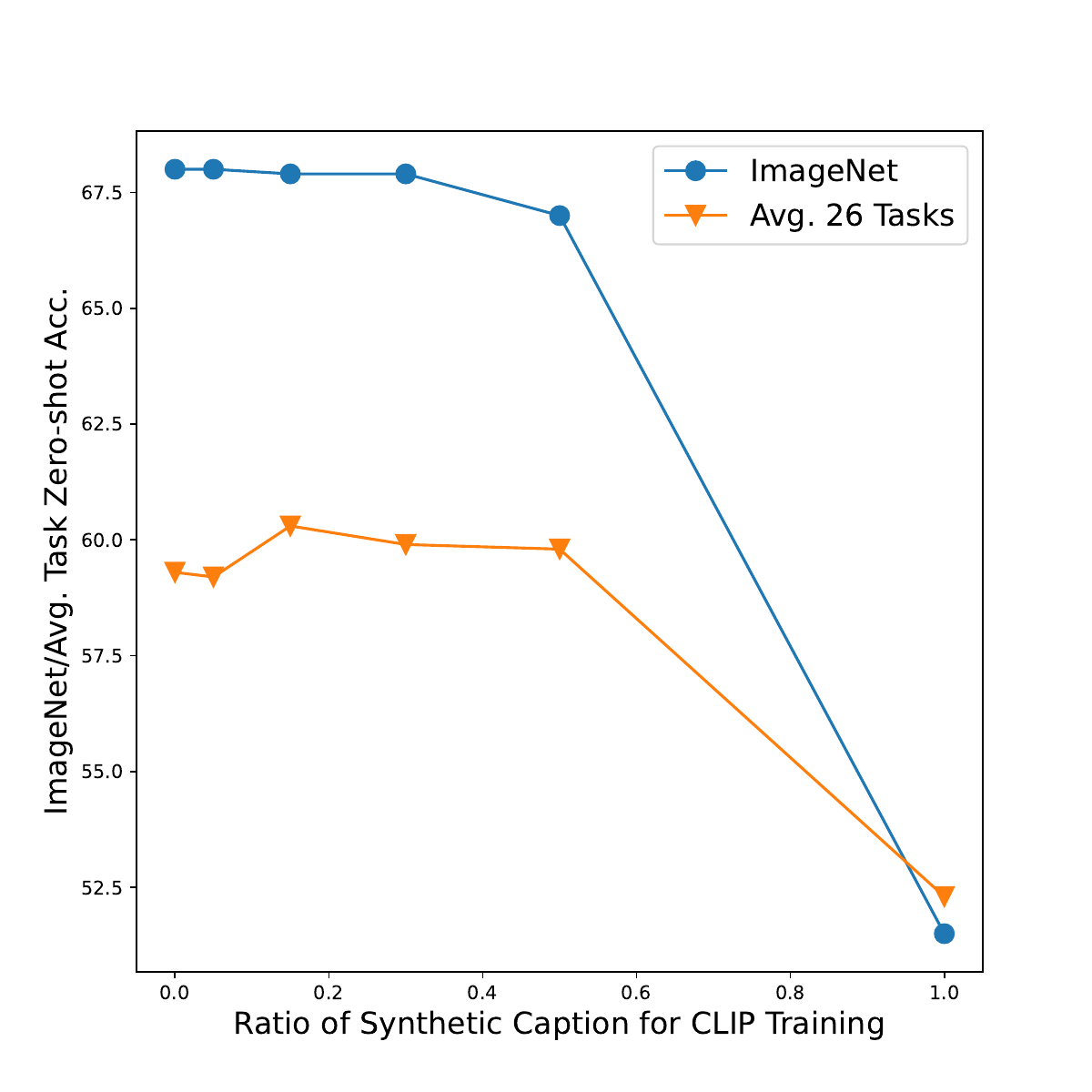}
\caption{Zero-shot classification accuracy on ImageNet and averaged 26 CLIP tasks with different ratio of mixing synthetic captions during training of various CLIP ViT-B/32 models.}
\label{fig:clip_mix}
\end{figure}

We ablate different mixing ratios of synthetic captions vs.~ImageNet zero-shot accuracy, and average accuracy across the 26 CLIP datasets in Fig.~\ref{fig:clip_mix} and notice that a high ratio of synthetic caption can reduce the performance significantly. A good trade-off ratio is around 15\%, which allows synthetic caption to complement alt-text, which is our default value throughout the paper.
This is likely due to two reasons: (i) human annotation optimizes alignment and is conservative on alt-texts when it concerns ambiguous image information.
For example, a ``\$18/night room'' in alt-texts could still supervise an image having a room of poor condition but is at risk of having mis-aligned description on price, so an annotator may remove that from alt-text; and (ii) existing benchmarks such as classification/retrieval test specific (object) classes instead of whole image alignment.

\section{Throughput of Different Text Decoders}
To scale captioner inference to billions of images, we ablate the throughput of different decoder setups in Table \ref{tbl:qps_GFLOPs}. We note that using such an LLM is 13.2\x slower than OPT (2.6 vs.~19.7 images per second).

\section{Hyperparameters}
\label{sec:hyperparams}
We detail the hyperparameters of the captioner in Table~\ref{tbl:hp_captioner}, downstream text-to-image training in Table~\ref{tbl:hp_dalle3} and CLIP training in Table~\ref{tbl:hp_clip}, respectively.

\subsection{Meta CLIP 1.2 Training}

Built upon the success of Meta CLIP, we have the following improvements in 1.2: 
\begin{itemize}
  \item \textbf{Fully Online Curation} that leverages data loader to transforms raw data distribution into training distribution. This allows for training on distribution instead of a concrete dataset and down sampling head distribution instead of one-time sampling. 
  \item \textbf{Scaling Batch Size} allows us to train Meta CLIP over 50B image-text pairs. We observe consistent improvement via 2$\times$ and 4$\times$ of batch size.
\end{itemize}

\section{Contributions}
\textbf{Hu Xu}: project lead, direction, prototyping of data quality and modeling, data sourcing and collection (download and multi-round mitigation), quality checking of multi-round  annotations, training and general evaluation, writing and code/model release.\\
\textbf{Po-Yao Huang}: downstream evaluation on text-to-image training and evaluation, general discussion on evaluation and presentation.\\
\textbf{Xiaoqing Ellen Tan}: multi-round data collection, annotation guideline, coordination with data vendors.\\
\textbf{Ching-Feng Yeh, Jacob Kahn, Christine Jou}: setup of human evaluation in multiple rounds, writing.\\
\textbf{Gargi Ghosh}: team management, GPU allocation.\\
\textbf{Omer Levy}: team management\\
\textbf{Luke Zettlemoyer}: direction, team coordination.\\
\textbf{Wen-tau Yih}: direction, discussion on annotation quality and evaluation.\\
\textbf{Shang-Wen Li}: team management/coordination, data mitigation, legal communication, data collection, data transfer, GPU allocation and public communication.\\
\textbf{Saining Xie}: advising on direction, writing and defining key terms and title.\\
\textbf{Christoph Feichtenhofer}: advising on experimental setups and presentation, multi-round writing and editing.

\begin{figure*}[!ht]
\begin{center}
\scalebox{0.9}
	    {
\fbox{\begin{minipage}[b]{\linewidth}
\begin{footnotesize}

\paragraph{Goal}
The goal of this task is to enhance the alignment in-between image and caption via caption editing, leveraging the metadata of the image (i.e. alt-text attributes). 
The collected data will be used to train a rewrite model for caption generation. The factoid knowledge and concrete visual concepts in alt-text is expected to be added to improve the caption and \textit{no extra} personal knowledge from annotators are expected as part of the improved caption. 
\newline

\paragraph{Task Description}
We provide a pair of \textit{(image, alt-text)} to annotators, and ask annotators to \textit{leverage the provided alt-text as factoid knowledge} and rewrite to improve the alignment between the \textit{caption} and the image. A better alignment means: 1) \textit{removing any nonfactual parts} in the caption; 2) \textit{adding missing information} into the caption (object shown in the image but not mentioned in caption). If the image-caption pair is 90\% aligned, make it 99\% aligned. \newline

\paragraph{Annotation Steps}
\begin{enumerate}[leftmargin=5mm,topsep=0pt,noitemsep]
    \item Copy and paste the ``Previous Caption'' to the box of ``Rewritten caption''.
    \item A concise starting prompt to describe what the image is about, such as ``a photo of'', ``a product photo of'', depends on types of images, rather than ``This image shows…''
    \item Use alt-text as much as possible if appropriate (mostly in 1st sentence) to improve the factuality of the caption.
    \begin{itemize}[leftmargin=3mm]
        \item 
        Paraphrasing is encouraged, but please do not change the meaning of the alt-text.
        \item Using concrete visual concepts in alt-texts as much as possible: write ``Bentley'' (alt-texts) as ``a Bentley'' instead of  ``a car''. 
        \item Alt-texts with metadata such as filenames/dates or ``photographed by ABC'' can be ignored.
        \item Using external tool (e.g., Google) is encouraged to help understand the alt-text.
    \end{itemize}
    \item Remove/Edit any hallucinated parts in the caption (anything that’s either not exists in the image or wrongly described, e.g., wrong color)
    \item Remove sentence describing theme/feeling of caption, e.g. ``overall this image gives an impression of xxx'' or imaginative description ``this boy must have a bright future.''. 
    \item To the extent the image contains people, please DO NOT provide any information about that person's
    \begin{itemize} 
        \item racial or ethnic origin (including skin color, hair color, apparent nationality or citizenship); 
        \item Sexual orientation; 
        \item Political affiliation;
        \item Health condition or disability;
        \item Religion;
        \item Membership in a Trade Union;
        \item Facial features, expression or emotion (e.g, smiling/crying as well as ``mood''), hair color (e.g., ``dark haired'', ``blonde-haired'', etc.);
        \item DO NOT add any identifying information about people or objects such as names, address and emails.
    \end{itemize}
    \item Add in visible missing details if there’s any. 
    
    \begin{itemize}[leftmargin=3mm]
        \item When less certain / in case of blurry image, use vague and general terms to describe the objects such as ``This may be NYC'' rather than ``This is NYC''; or ``animal'' instead of ``dog''/``cat'' (when it's hard to judge detailed type).
        \item Transcribe any readable characters in the image.
    \end{itemize}
    \item Check the overall structure (deductive structure etc) of the rewritten caption.
    \begin{itemize}[leftmargin=3mm]
        \item Make sure everything in the caption is factual.
        \item Check the structure of caption (see the next section). 
    \end{itemize}
\end{enumerate}

\paragraph{Structure of Caption}
\begin{enumerate}[leftmargin=5mm,topsep=0pt,noitemsep]
    \item Caption structure
    \begin{itemize}[leftmargin=3mm]
        \item Objects: A good dense caption should follow a ``deductive structure'' where it typically starts with a general statement, followed by subjects, secondary objects, background, and concluding with minor details. 
        \item Order of objects: Similar to how a human would usually read images e.g., ``left to right'', ``top to bottom'', or ``near to far'' order. Once done with describing the most salient objects, for secondary objects and backgrounds that are hard to sort by saliency, we can arrange secondary objects and background elements in a similar way, depending on the image structure.
        \begin{itemize}
            \item The default spatial terms is based on viewer’s angle (3rd person); if 1st person view angle is needed, explicitly write down that angle: ``on her left is a cute dog'';
            \item Describe spatial relation from big to small, from main to accessory: '' ... a cake. There're 4 cherries on the cake.''.  
            \item Count objects of the same type when it is less than or equal to 10; for more than 10 objects, annotator may use the word ``many x''.
        \end{itemize}
        \item Long paragraph: Please split a long paragraph into shorter and coherent paragraphs, and organize them with a clear logical order for easier understanding. 
    \end{itemize}
    \item Caption length
    \begin{itemize}[leftmargin=3mm]
        \item Conciseness, correlates with “complexity” of the image. Though we want to have detailed descriptions, we also want to have the details being described in a concise way. If there is only one object present in the image, we shouldn’t have a long paragraph.
    \end{itemize}
\end{enumerate}
\end{footnotesize}
\end{minipage}}
}
\caption{Annotation guideline.}
\label{fig:annotation_instruct}
\end{center}
\end{figure*}

\begin{table*}[ht]
\centering
\tablestyle{1.5pt}{1.1}
\begin{tabularx}{0.99\linewidth}{c | p{0.2\linewidth} | p{0.25\linewidth} | p{0.25\linewidth} }
Image & Alt-Text (Round 1) & Round 2 & Round 3\\
\shline
\raisebox{-1.\height}{\includegraphics[width=0.2\textwidth]{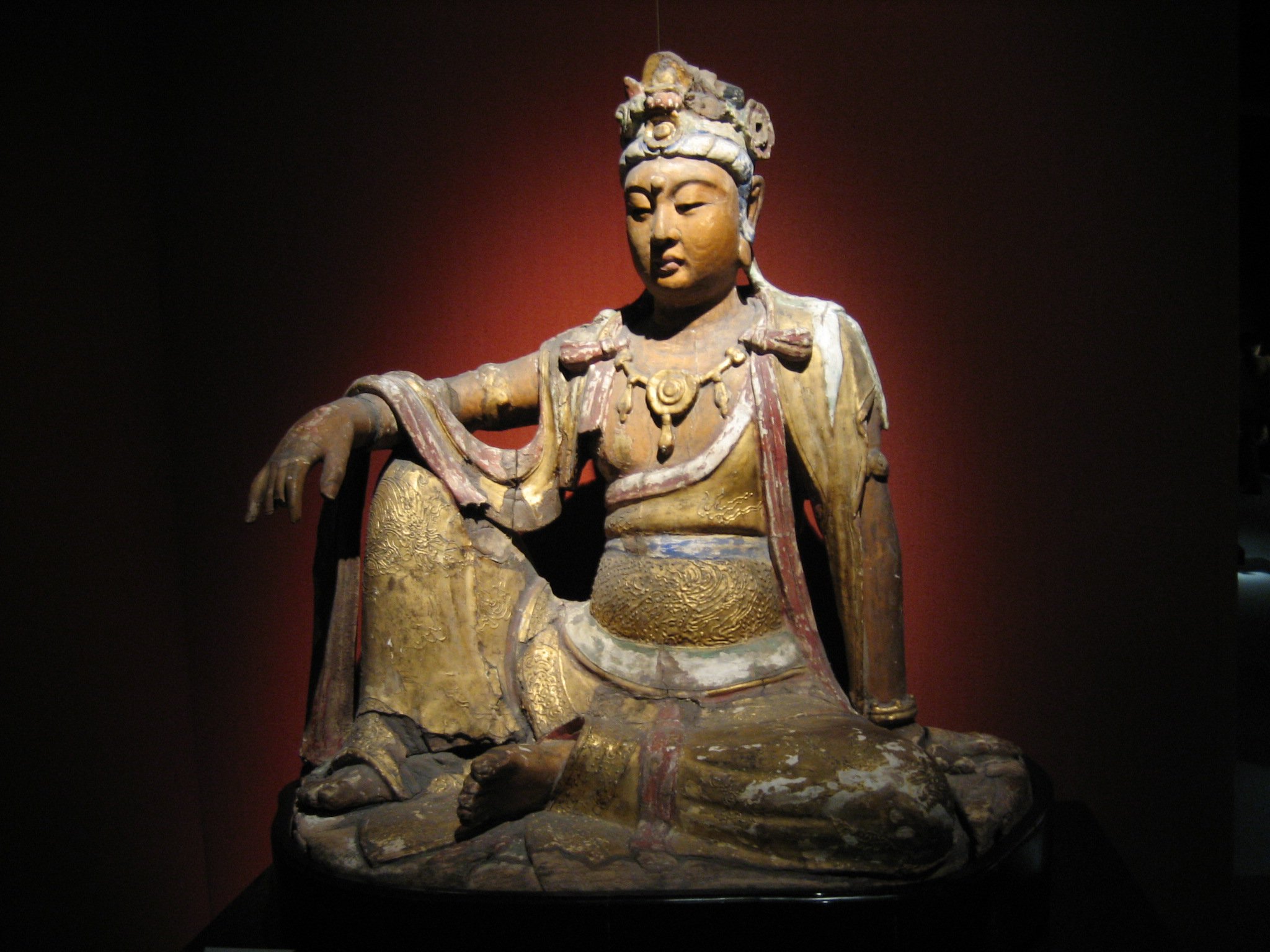}} & A wooden gilded statue of the bodhisattva Avalokiteśvara from the Chinese Song Dynasty (960-1279), from the Shanghai Museum. Français: Statue en bois doré de bodhisattva Avalokiteśvara. Dynastie Song (960-1279), Musée de Shanghai. & A photograph of a {\color{blue}wooden gilded statue of the bodhisattva Avalokiteśvara from the Chinese Song Dynasty}. The image depicts a statue of a seated figure placed on a black pedestal. The figure is adorned with elaborate attire and jewelry, suggesting it might represent royalty or divinity. The clothing and accessories feature intricate designs and patterns. Unfortunately, the face of the statue is obscured by a beige-colored rectangle. The background has a dark red hue, creating contrast with the lighter color of the statue. & A photo of a {\color{blue}wooden gilded statue of the bodhisattva Avalokiteśvara from the Chinese Song Dynasty  (960-1279), from the Shanghai Museum}.  The image depicts a statue of a seated figure placed on a black pedestal. The figure is adorned with elaborate attire and jewelry, suggesting it might represent royalty or divinity. The clothing and accessories feature intricate designs and patterns. Unfortunately, the face of the statue is obscured by a beige-colored rectangle. The background has a dark red hue, creating contrast with the lighter color of the statue.\\
\hline
\raisebox{-1.\height}{\includegraphics[width=0.2\textwidth]{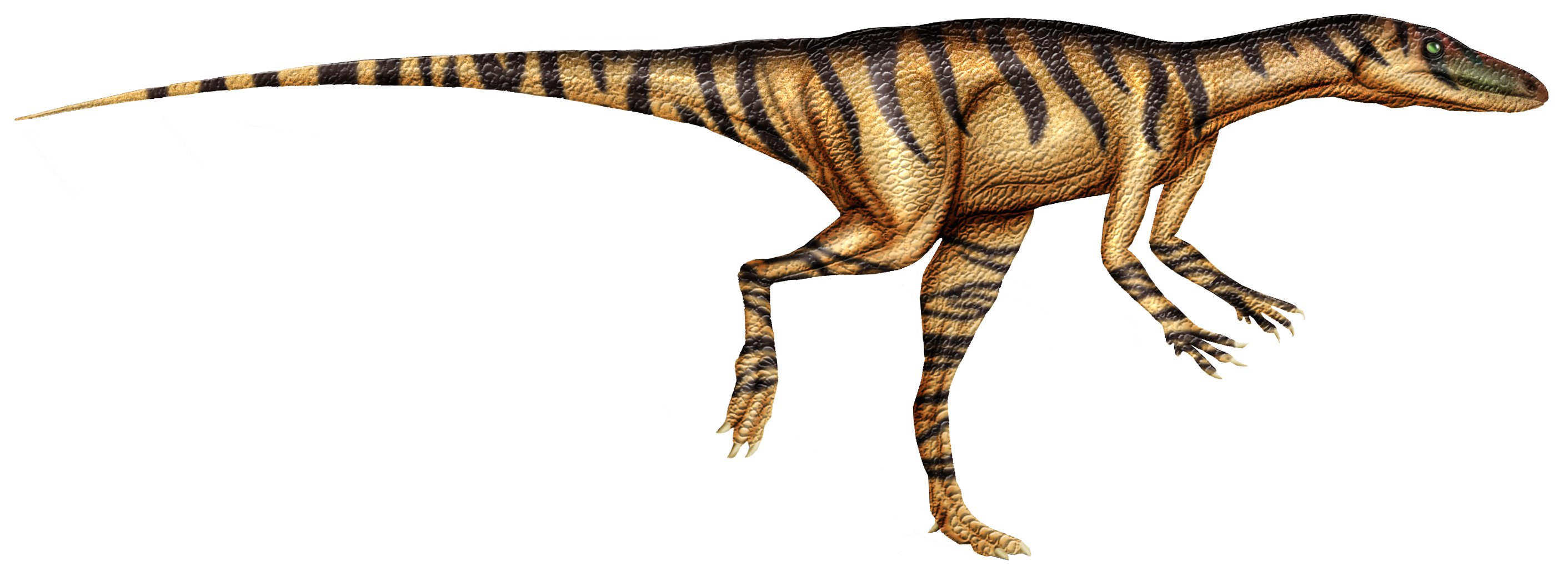}} & Alwalkeria Illustration of Alwalkeria maleriensis by Karkemish & A 3D render of a dinosaur, specifically an {\color{blue}Alwalkeria}. It is depicted in a running pose, with its left foot forward and its right foot back. The dinosaur's body is a mix of orange, yellow, and brown colors, with black stripes on its back and tail. It has a long, curved tail and sharp teeth. It is isolated in a white background. & A 3D render of a dinosaur, specifically an {\color{blue}Alwalkeria}. It is depicted in a running pose, with its left foot forward and its right foot back. The dinosaur's body is a mix of orange, yellow, and brown colors, with black stripes on its back and tail. It has a long, curved tail and sharp teeth. It is isolated on a white background.\\
\hline
\raisebox{-1\height}{\includegraphics[width=0.2\textwidth]{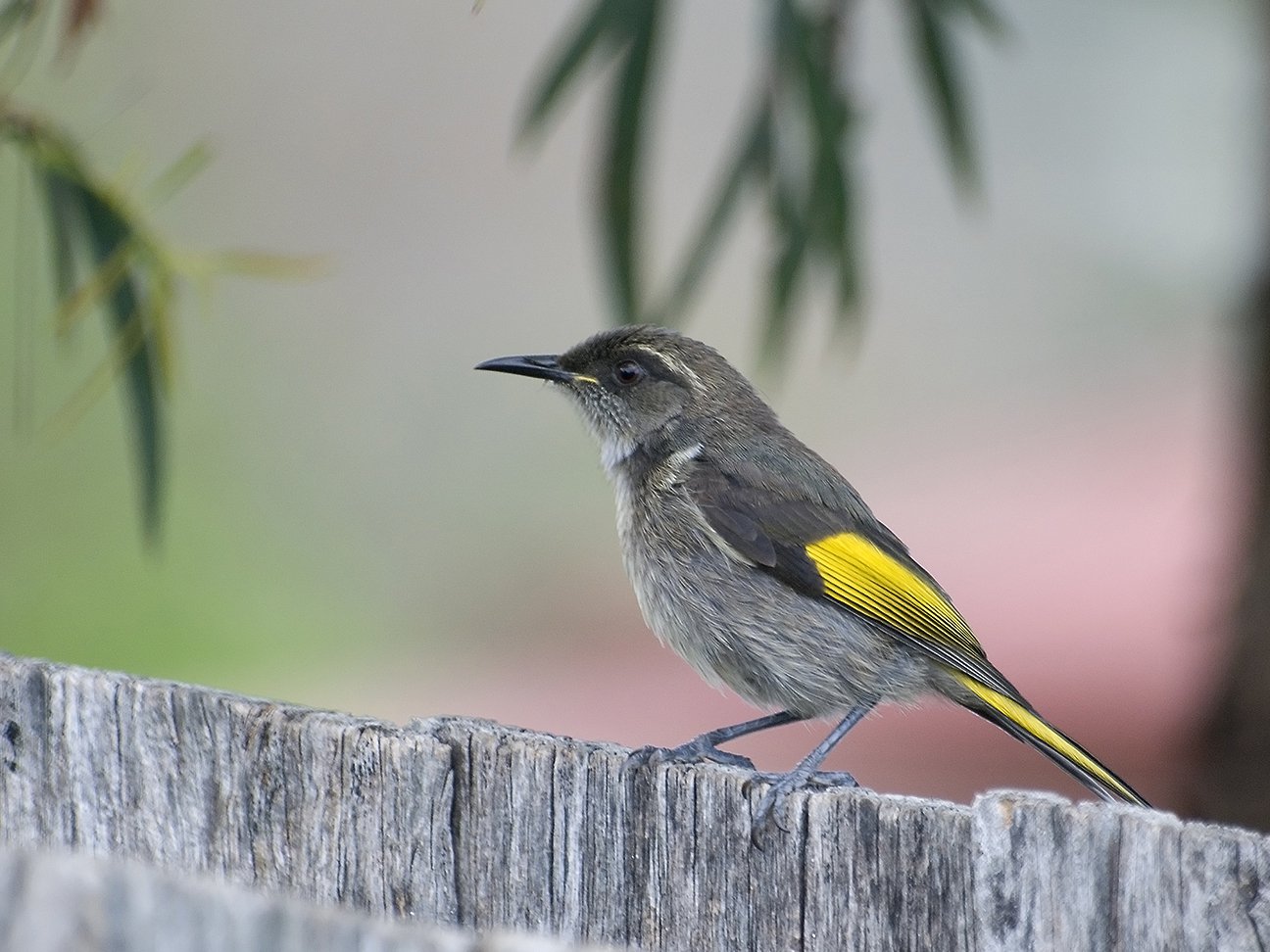}} & Juvenile male in Tasmania, Australia A male crescent honeyeater (Phylidonyris pyrrhopterus) in Lindisfarne, Tasmania, Australia & A close-up photo of a {\color{blue}juvenile male crescent honeyeater} bird perched on a gray picket fence. The bird has a dark grey plumage, pale gray underparts, and yellow wing patches. The bird also has a black bill, a pair of dark eyes, and gray legs. A few leaves and branches of a tree are partly visible in the blurring background. & A photo of a {\color{blue}juvenile male crescent honeyeater, (Phylidonyris pyrrhopterus)}. The bird is perched on a gray picket fence in {\color{blue}Lindisfarne, Tasmania, Australia}. The bird has a dark grey plumage, pale gray underparts, and yellow wing patches. The bird also has a black bill, a pair of dark eyes, and gray legs. The background has a few leaves and branches of a tree that are blurred.\\
\hline
\end{tabularx}
\caption{Examples of annotated images from WIT dataset in multi-rounds.}
\vspace{5pt}
\label{tbl:sbs_annotation_wit}
\end{table*}

\begin{table*}[ht]
\centering
\tablestyle{1.5pt}{1.1}
\begin{tabularx}{0.99\linewidth}{p{0.2\linewidth} | p{0.38\linewidth} | p{0.38\linewidth} }
Alt-Text (Round 1) & Round 2 & Round 3\\
\shline
Vintage 1992 University of Miami Hurricanes Orange Bowl T-shirt & A photo of a green t-shirt with a straight hem. The t-shirt is made of a thick, high-quality fabric that is soft to the touch. The color is a deep green, almost a forest green, with a slight sheen to it. The image is taken on a yellow background. & A photo of a {\color{blue} Vintage 1992 University of Miami Hurricanes Orange Bowl} T-shirt. The T-shirt is made of a thick fabric. The color of the T-shirt is a deep green, almost a forest green color. The t-shirt is depicted on a yellow background.\\
\hline
Aqua Recessed swimming Pool 11x11cm MR 16 IP68 Stainless Steel AISI 316 & A product photo of {\color{blue} Aqua Recessed swimming Pool 11x11cm MR 16 IP68 Stainless Steel AISI 316 light fixture}. The image shows a round, stainless steel submersible outdoor pool light fixture with a flat, glass lens. The light is recessed into the fixture and surrounded by a ring of four metal flanges. The flanges have small holes drilled in them. The light fixture is secured to the ground with a large bolt in the center. The light source is not visible, but it appears to be an LED or other small light source. The image is on a white background, and the light fixture is the only object in the image. & A product photo of {\color{blue} Aqua Recessed swimming Pool 11x11cm MR 16 IP68 Stainless Steel AISI 316 light fixture}. The image shows a round, stainless steel submersible outdoor pool light fixture with a flat, glass lens. The light is recessed into the fixture and surrounded by a ring of four metal flanges. The flanges have small holes drilled in them. The light fixture is secured to the ground with a large bolt in the center. The light source is not visible, but it appears to be an LED or other small light source. The image is on a white background, and the light fixture is the only object in the image.\\
\hline
North Carolina Tar Heels Team Logo Gray Adjustable Hat GS & a product photo of a {\color{blue} North Carolina Tar Heels Gray Adjustable Hat GS}. The hat is a gray and blue snapback hat with a blue logo of interlocking letters ``NC'' on the front. The hat has a blue flat bill and a blue adjustable snapback closure on the back. The logo is surrounded by a white outline, creating a sharp contrast with the gray background. The image consists of two photos of the same hat, a smaller one in the top left section that shows the back of the hat, and a bigger one in the bottom right section showing the front of the hat. The background of the image is white. & A product photo of a {\color{blue} North Carolina Tar Heels Gray Adjustable Hat GS}. The hat is a gray and blue snapback hat with a blue logo of interlocking letters ``NC'' on the front. The hat has a blue flat bill that contains a label sticker that is hard to see, and a blue adjustable snapback closure on the back. The logo is surrounded by a white outline, creating a sharp contrast with the gray background. The image consists of two photos of the same hat, a smaller one in the top left section that shows the back of the hat, and a bigger one in the bottom right section showing the front of the hat. The background of the image is white.\\
\hline
Data Visualization with Python and Matplotlib & A photo of image features a graph created using {\color{blue} Matplotlib, a widely-used data visualization library for Python}. The graph showcases three circles arranged in a spiral-like pattern. The innermost circle contains two distinct -shaped images in yellow and blue, while a quarter shape is prominently orange in color. Across the image is the text ``Matplotlib''. The entire composition is set against a grey background. & A photo of image representing {\color{blue} data visualization using Python and Matplotlib}. The image showcases three circles arranged in a spiral-like pattern. The innermost circle contains two distinct -shaped images in yellow and blue, while a quarter shape is prominently orange in color. Across the image is the text ``Matplotlib''. The entire composition is set against a grey background.\\
\hline
\end{tabularx}
\caption{Re-aligned alt-texts from MetaCLIP~\cite{metaclip} images.}
\label{tbl:sbs_annotation_metaclip}
\end{table*}

\end{document}